\documentclass{article}

 \usepackage[main, final]{neurips_2026}

\usepackage[utf8]{inputenc} 
\usepackage[T1]{fontenc}    
\usepackage{hyperref}       
\usepackage{url}            
\usepackage{booktabs}       
\usepackage{amsfonts}       
\usepackage{nicefrac}       
\usepackage{microtype}      
\usepackage{xcolor}         

\usepackage{amsmath}
\usepackage{amssymb}
\usepackage{mathtools}
\usepackage{amsthm}
\usepackage{algorithm}
\usepackage{algorithmic}

\usepackage{multicol}
\usepackage[table]{xcolor}
\usepackage{listings}
\usepackage{placeins}
\usepackage{makecell}
\usepackage{subcaption}

\usepackage{microtype}
\usepackage{graphicx}
\usepackage{booktabs} 
\usepackage{multirow}

\usepackage{caption}
\title{Order Matters: Competition-Guided Query Ordering for RNN-Based Object Detection}

\author{%
  Shengjian Wu\textsuperscript{1,2}
  \qquad
  Li Sun\textsuperscript{1,3}\thanks{%
    Correspondence to: Li Sun (\texttt{sunli@ee.ecnu.edu.cn}).\enspace
    \textsuperscript{1}\,Shanghai Key Laboratory of Multidimensional Information
    Processing, East China Normal University;
    \textsuperscript{2}\,FinVolution Group;
    \textsuperscript{3}\,Key Laboratory of Advanced Theory and Application in
    Statistics and Data Science, East China Normal University.}
  \qquad
  Yu Shangguan\textsuperscript{1}
  \qquad
  Qingli Li\textsuperscript{1}
}

\begin{document}

\maketitle

\begin{abstract}
DETR-style detectors use one-to-one bipartite matching during training to assign object queries to ground-truth objects, enabling end-to-end set prediction without non-maximum suppression (NMS).
However, without an explicit de-duplication procedure, multiple queries can still produce highly similar hypotheses for the same object, making training unstable and predictions less decisive.
Inspired by the sequential ordering of NMS, we propose DETRNN, a plug-and-play module that turns unordered object queries into a competition-aware sequence for recurrent refinement.
DETRNN builds an explicit confidence-and-similarity based order from prior predictions, then refines queries with an RNN along this order to model competition inside the decoder.
This ordered recurrent refinement reduces redundant predictions, stabilizes optimization, and improves final detection accuracy.
Experiments on multiple DETR-style detectors show consistent gains with comparable efficiency.
Code repository: \url{https://github.com/nigelwoo2008-coder/DETRNN}.
\end{abstract}

\section{Introduction}
\begin{figure}[t]
    \centering
    \makebox[\linewidth][c]{
  \begin{minipage}{0.65\linewidth}
    \begin{subfigure}[t]{0.33\linewidth}
      \centering
      \includegraphics[width=\linewidth]{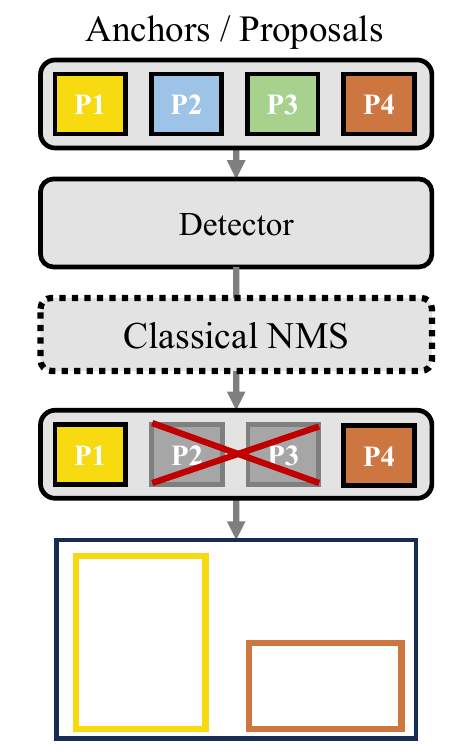}
      \caption{NMS}
    \end{subfigure}\hfill
    \begin{subfigure}[t]{0.33\linewidth}
      \centering
      \includegraphics[width=\linewidth]{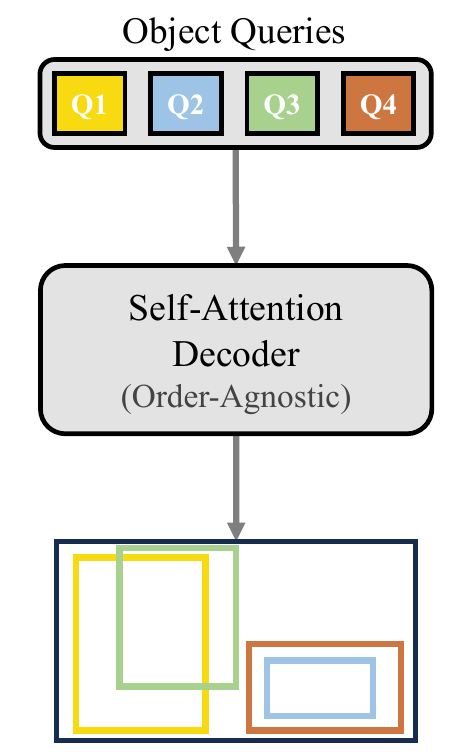}
      \caption{DETR}
    \end{subfigure}\hfill
    \begin{subfigure}[t]{0.33\linewidth}
      \centering
      \includegraphics[width=\linewidth]{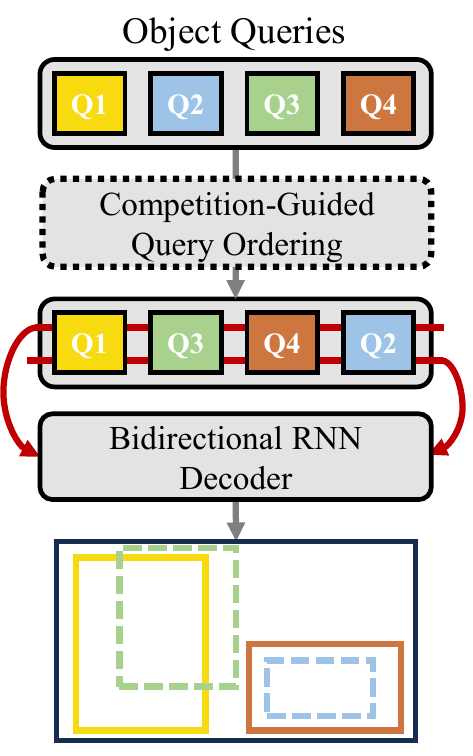}
      \caption{DETRNN}
    \end{subfigure}
  \end{minipage}
}
\caption{\textbf{A comparison of de-duplication mechanisms.} Classical NMS de-duplicates by removing highly overlapping proposals with a fixed IoU threshold (a). 
DETR predicts end-to-end, but its order-agnostic decoder can still yield duplicated results (b).
DETRNN internalizes de-duplication into the decoder: CGQO builds a competition-aware query order, and a Bi-RNN refines queries sequentially to softly suppress redundant predictions (dashed boxes) without discarding outputs (c).}
\label{fig:teaser}
  \vspace{-0.3 cm}
\end{figure}

Transformer-based detectors such as DETR~\cite{carion2020end} unify object detection into an end-to-end set prediction problem.
During training, DETR applies one-to-one bipartite matching between object queries and ground-truth objects, which removes post-processing NMS, avoiding manual threshold tuning.
Despite this elegant formulation, DETR still suffers from redundant hypotheses, especially in dense scenes (Fig.~\ref{fig:teaser}b).
Multiple queries can converge to similar regions with close confidence scores, but only one prediction can be matched to each ground-truth object, leaving the others as competing duplicates and slowing convergence.

A useful reference point is classical NMS.
NMS resolves redundancy via an explicit sequential procedure: it ranks predictions and greedily removes highly overlapping candidates (Fig.~\ref{fig:teaser}a).
This sequential order is a strong and simple cue for organizing competition among duplicates.
However, using NMS as a hard post-processing step is not always ideal: it is non-learnable, can be threshold-sensitive, and may even hurt in some cases.
This motivates a different goal: rather than performing NMS as post-processing, we use an explicit competition order to guide learnable decoder refinement and reduce redundancy.

In DETR, query interaction is mainly handled by self-attention.
Recent analyses (e.g., DAC-DETR~\cite{hu2023dac}, OneNet~\cite{peize2020onenet}) suggest that self-attention, together with one-to-one matching used during training, plays an NMS-like role by suppressing duplicates over decoders.
Yet self-attention is parallel and order-agnostic, and it does not provide an explicit competition order.
Therefore, prior work typically injects competition priors by modifying attention patterns or biases, such as easing winner-versus-competitor interaction in EASE-DETR~\cite{gao2024EASEDETRET} or introducing relation priors in Relation-DETR~\cite{hou2024relation}, or by applying decoder-wise NMS to discard redundant queries, e.g., DDQ~\cite{Zhang_2023_CVPR}.
While effective, these designs remain within parallel attention and rely on carefully crafted attention modifications or query pruning.

We propose a more direct alternative: use an RNN to refine queries along an explicit competition sequence.
RNNs are naturally suited for sequential modeling, but DETR queries have an arbitrary order with no competition cue.
To bridge this gap, we introduce \textbf{DETRNN}, which constructs a competition sequence from prior predictions
 by sorting queries using confidence and similarity under \textbf{Competition-Guided Query Ordering} (CGQO), mimicking the ranking-and-grouping behavior of NMS.
A learnable RNN-based query interaction then refines queries along this sequence, making redundant competitors easier to suppress  (Fig.~\ref{fig:teaser}c).
As a result, DETRNN improves training stability and detection quality by suppressing duplicates and making the correct prediction stand out.

To our knowledge, DETRNN is among the first attempts to use an RNN as the query-interaction module in DETR decoders, enabling redundant predictions to be suppressed during decoding. We validate DETRNN on diverse DETR-family baselines, including Def-DETR++~\cite{zhu2020deformable}, DINO~\cite{zhang2022dino}, MS-DETR~\cite{Zhao2024MSDETRED}, and recent LoRA-DETR~\cite{zhang2026integrating}. The consistent experimental gains across these architectures show a practical accuracy--efficiency trade-off.

The contributions of this paper are summarized as follows:

\begin{itemize}
\item We identify query ordering as a missing inductive bias for de-duplication in end-to-end detection, turning redundancy suppression into an explicit sequential refinement problem.

\item We introduce a plug-and-play ordered recurrent query interaction: CGQO builds a competition order from prediction similarity, and an RNN refines object queries along this order to suppress redundant hypotheses.

\item Extensive experiments validate CGQO-guided RNN interaction as an effective alternative to decoder self-attention, reducing prediction redundancy and improving AP across diverse detectors with comparable efficiency.
\end{itemize}

\section{Related Work}

\subsection{End-to-end Object Detectors}
DETR~\cite{carion2020end} introduced a transformer-based end-to-end detector, but its high computational cost, slow convergence, and weak small-object performance limit practical use.
Def-DETR~\cite{zhu2020deformable} alleviates these issues with deformable attention.
Other query-based detectors, such as Sparse R-CNN~\cite{Sun2020SparseRE} and AdaMixer~\cite{gao2022AdaMixerAF}, redesign query decoding with learnable proposals, dynamic convolution, or RoI-based feature extraction.
Another line improves convergence and matching stability by enriching query supervision.
DN-DETR~\cite{li2022dn} and DINO~\cite{zhang2022dino} introduce denoising queries, while Hybrid-DETR~\cite{jia2022detrs}, MS-DETR~\cite{Zhao2024MSDETRED}, LoRA-DETR~\cite{zhang2026integrating} and related methods~\cite{zong2023detrs,chen2022group,huang2024deim,zhang2024mr,wu2024ODDETROD,liu2023detection} expand query groups or use richer GT--prediction matching to provide more positive supervision.
Beyond supervision and matching, stage-wise reuse~\cite{Chen2022EnhancedTO,zhao2023recursivedet}, ranking-aware~\cite{Pu2023RankDETR}, and dynamic-query methods~\cite{Dai2021DynamicDE,Kang2026PaQDETRLP,huang2024dq} improve query propagation, quality, or allocation.
These methods improve optimization or query utilization, but they do not explicitly model the order of competition among redundant queries.

OneNet~\cite{peize2020onenet} points out that NMS-free detection relies on one-to-one supervision to avoid duplicate predictions.
DAC-DETR~\cite{hu2023dac} further shows that decoder self-attention plays a key role in redundancy removal by mediating query--query interaction.
However, when decoding hundreds of object queries, parallel self-attention may still struggle to focus on close competitors, often requiring many epochs and multiple decoder stages.
Recent methods therefore improve de-duplication in two main ways.
Some select or prune queries, e.g., filtering salient queries before decoding~\cite{hou2024salience,cao2025dsdet} or applying NMS-style query pruning across decoder stages~\cite{Zhang_2023_CVPR,OuyangZhang2022NMSSB}.
Others reshape parallel query interaction with spatial/IoU or prediction-aware attention priors~\cite{Zhao2022IoUEnhancedAF,gao2022AdaMixerAF}, and recent competition-aware designs such as Relation-DETR~\cite{hou2024relation}, EASE-DETR~\cite{gao2024EASEDETRET}, and Dual-R-DETR (formerly Route-DETR)~\cite{Zhang2025DualRDETRRQ}.
Rather than further modifying parallel pairwise attention or pruning queries, we study ordered recurrent query interaction, where an RNN exploits an explicit competition order to suppress redundant hypotheses inside the decoder.

\subsection{Vision RNNs}
Recurrent Neural Networks (RNNs), including LSTM~\cite{hochreiter1997long} and GRU~\cite{chung2014empirical}, were originally designed for temporal modeling, and recent variants such as Mamba~\cite{Gu2023MambaLS}, RWKV~\cite{peng2023RWKVRR}, TTT~\cite{sun2024LearningT}, and xLSTM~\cite{beck2024xLSTMEL} further improve efficient long-range sequence modeling.
In vision, RNN-style models are often used to process image tokens or enhance feature extractors: Vision Mamba~\cite{Zhu2024VisionME} and Vision-RWKV~\cite{Duan2024VisionRWKVEA} model image tokens sequentially, VMamba~\cite{Liu2024VMambaVS} explores multidirectional scanning, and Mamba-YOLO~\cite{Wang2024MambaYS} integrates SSM blocks into ViTs and FPN.
These methods mainly improve visual representations, while DETR's redundancy handling depends on decoder query interaction.
A related detector, Recurrent DETR~\cite{choi2023recurrent}, iterates transformer decoder blocks for variable-size prediction, rather than using RNNs for query interaction.
In contrast, DETRNN applies recurrent modeling inside the DETR decoder to refine competition-ordered queries and reduce redundancy.

\section{Method}
\label{sec:mthd}
\subsection{Preliminaries}

\noindent \textbf{DETR}-style detectors employ a fixed set of learnable object queries
$\mathbf{Z}=\{z_1,z_2,\cdots,z_N\}\in\mathbb{R}^{N\times d}$ to produce a set of detection predictions,
where each query $z_i$ outputs a classification score and a bounding box.
Hungarian bipartite matching enforces a one-to-one assignment between predictions and ground truth objects,
enabling end-to-end detection without explicit NMS.

In practice, however, multiple queries may converge to highly overlapping hypotheses around the same object with similar confidence,
creating redundant competitors.
Since only one prediction can be matched to each ground truth, small score fluctuations may switch the matched prediction during training,
making supervision noisy and slowing convergence, especially in crowded scenes.

To mitigate this issue, decoder self-attention is used for query--query interaction.
Recent analyses (DAC-DETR~\cite{hu2023dac}, OneNet~\cite{peize2020onenet}) suggest that self-attention, together with one-to-one matching,
implicitly plays an NMS-like role by gradually suppressing duplicates.
Nevertheless, self-attention remains parallel and order-agnostic, and does not explicitly organize competitors
into a sequential competition procedure.

\noindent \textbf{Recurrent Neural Networks (RNNs)} are designed to process sequential data by maintaining hidden states across time steps. Given a sequence ${x_1, x_2, ..., x_T}$, a standard RNN, denoted as $[o_t,h_t]=\text{RNN}(x_t, h_{t-1})$, computes the hidden state $h_t$ and output $o_t$ at time step $t$ as:
\begin{equation}\label{eq:eq2} 
\begin{aligned}h_t = f(\mathbf{W}_h h_{t-1} + \mathbf{W}_x x_t + b) \quad o_t=\mathbf{W}_o h_t \end{aligned}    
\end{equation}
where $f(\cdot)$ is a non-linear activation function, e.g., tanh; $\mathbf{W}_x$, $\mathbf{W}_h$ and $\mathbf{W}_o$ are static weight matrices for the input, the hidden state transition and the output, respectively, with $b$ as a bias term.

RNN-style models provide a natural sequential inductive bias by refining representations along an ordered stream.
This makes them a suitable candidate for modeling competition in set prediction, provided that the query sequence encodes meaningful competition cues. Recent advances in RNNs integrate transformer-inspired mechanisms that dynamically adjust weight matrices $\mathbf{W}$ based on the current input $x_t$, enhancing memory and flexibility.

\begin{figure}[t]
\centering
\hspace{0.05 cm}
\includegraphics[width=13.8cm]{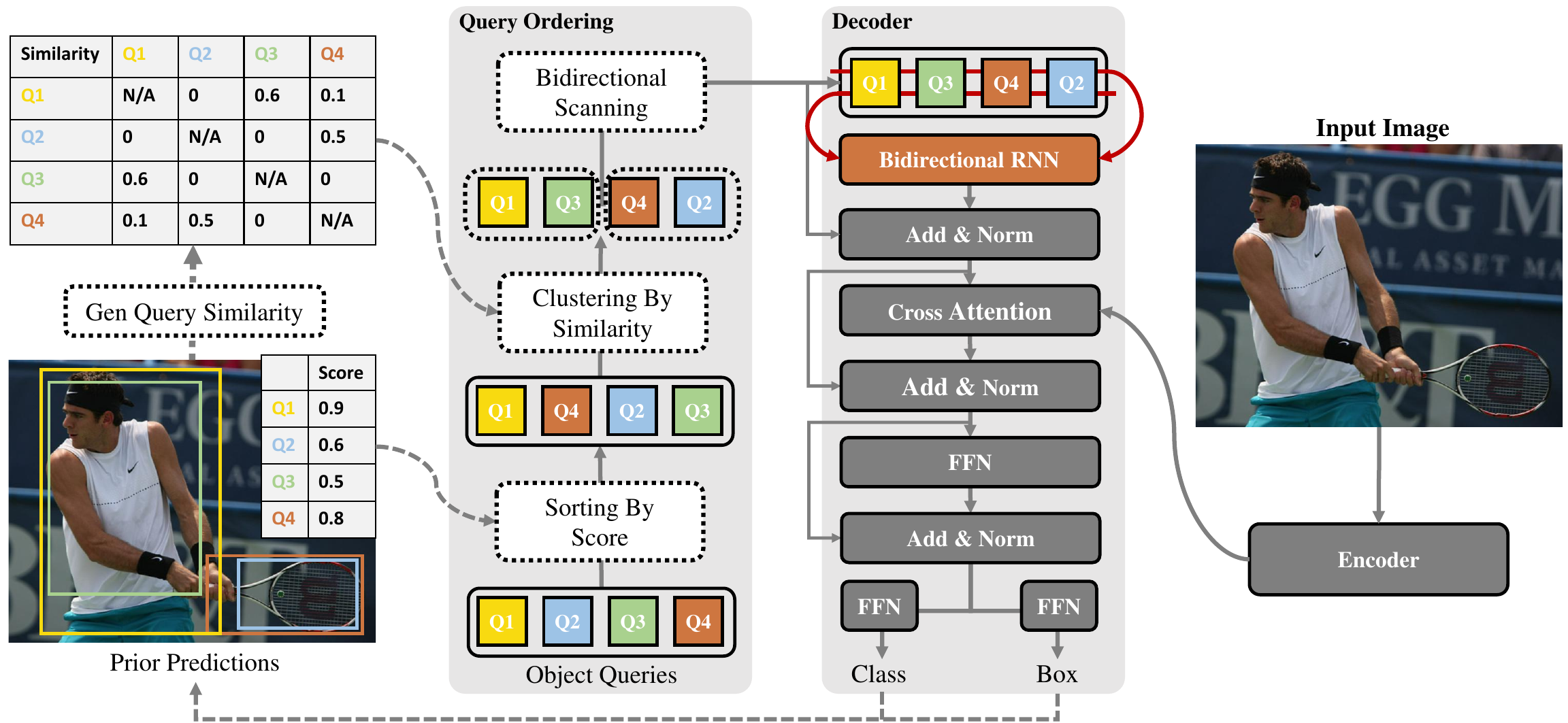}
\caption{
Overview of DETRNN.
Given prior predictions from encoder proposals or preceding decoder outputs, Competition-Guided Query Ordering (CGQO) first constructs an explicit competition order among queries.
Specifically, queries are ranked by confidence; the top-scoring query is selected as a pivot, and other queries are grouped and reordered by their similarity to the pivot (highly similar competitors are placed nearby).
This pivot-and-grouping process is repeated until all queries are ordered, yielding a competition-aware sequence. Bidirectional RNN then replaces decoder self-attention and refines queries sequentially along this order, enabling learnable competition resolution inside the decoder.
Unlike NMS, CGQO does not discard queries; it only provides ordering cues that the decoder can internalize for de-duplication.
}
\label{fig:pipeline}
\end{figure}

\subsection{Architecture Overview}
\label{subsec:method_overview}
Building on the observation above, we aim to make query competition resolvable through an explicit order and learnable sequential refinement inside the decoder.

A useful reference point is classical NMS.
Beyond being a post-processing rule, its procedure exposes an effective de-duplication ordering cue: it ranks candidates by confidence and organizes close duplicates to follow top-scoring ones.
We aim to internalize this ordering cue into the decoder as a learnable refinement process, rather than applying threshold-sensitive post-processing.
However, decoder self-attention is parallel and order-agnostic, and does not provide an explicit competition order to exploit.

To this end, we propose DETRNN, a plug-and-play sequential refinement module composed of two steps (Fig.~\ref{fig:pipeline}).
First, Competition-Guided Query Ordering (CGQO) converts the unordered query set into an explicit competition-ordered stream based on prior predictions.
Then, an RNN-based interaction refines queries sequentially along this order, suppressing redundant competitors more effectively.

Overall, DETRNN turns de-duplication into an explicit sequential refinement process inside the decoder.
We next detail Competition-Guided Query Ordering (CGQO) and the Bi-RNN refinement module.

\begin{algorithm}[t]
\small
\renewcommand{\algorithmiccomment}[1]{\hfill{\color{gray}\small$\triangleright$~#1}}
\caption{Competition-Guided Query Ordering}
\label{alg:cgqo}

\begin{algorithmic}
\STATE \textbf{Require:}
\STATE \hspace{0.5em} $b \in \mathbb{R}^{B \times N \times 4}$ (bounding boxes),
$s \in \mathbb{R}^{B \times N \times (C)}$ (class probabilities),
\STATE \hspace{0.5em} $c \in \mathbb{R}^{B \times N}$ (confidence scores, $c=\max(s)$),
$z \in \mathbb{R}^{B \times N \times d}$ (object queries),
\STATE \hspace{0.5em} $T \in \mathbb{R}^{B \times N}$ (pivot-wise similarity thresholds),
$\alpha$ (IoU/cosine weight)

\STATE \textbf{Definitions:}
\STATE \hspace{0.5em} $all\_sim \gets (1-\alpha)\,\mathrm{IoU}(b) + \alpha\,\mathrm{cosine}(s)$
\algorithmiccomment{$all\_sim \in \mathbb{R}^{B \times N \times N}$}
\end{algorithmic}

\vspace{-0.3em}

\begin{algorithmic}[1]
\STATE $reorder\_queries \gets \emptyset$
\FOR{each batch index $i = 1$ to $B$}
    \STATE $order \gets \mathrm{argsort}(c[i])$ descending
    \algorithmiccomment{rank by confidence}
    \STATE $clusters \gets \emptyset$
    \WHILE{$order \neq \emptyset$}
        \STATE $top \gets order[0], \quad \tau \gets T[i, top]$
        \algorithmiccomment{current pivot and its pivot-specific threshold}
        \STATE $sim \gets all\_sim[i][top,\, order]$
        \STATE $inds \gets order[sim \ge \tau]$
        \algorithmiccomment{group close competitors and get indices of them}
        \STATE Append $inds$ to $clusters$
        \STATE Remove $inds$ from $order$
        \algorithmiccomment{remove grouped queries}
    \ENDWHILE
    \STATE $ind \gets$ concatenate all $clusters$
    \STATE Append $z[i][ind]$ to $reorder\_queries$
\ENDFOR
\STATE \textbf{return} $reorder\_queries$
\vspace{-0.1cm}
\end{algorithmic}
\end{algorithm}

\subsection{Competition-Guided Query Ordering}
\label{subsec:nms_sort}
CGQO is the first step of DETRNN: it converts the unordered query set into an explicit competition-ordered stream using prior predictions (encoder proposals or preceding decoder outputs) (Fig.~\ref{fig:pipeline}).
Unlike post-processing NMS, CGQO keeps \emph{all} queries and only exposes an ordering cue for the decoder to learn de-duplication end-to-end.
In practice, we use a pivot-specific threshold to adapt grouping to local crowding; the rule is given in the paragraph \emph{Density-adaptive threshold}.

\vspace{-0.2cm}
\paragraph{Sorting by confidence and similarity.}
CGQO follows a simple pivot-and-group procedure. For each image, we rank queries by confidence and take the top-scoring one as a pivot $z_{\text{ref}}$ with box $b_{\text{ref}}$ and score vector $s_{\text{ref}}\in\mathbb{R}^{C}$.
For every remaining query $z_i$ (with $b_i$ and $s_i$), we compute its similarity to the pivot by combining geometric overlap and semantic agreement:
\begin{equation}\label{eq:sim}
\mathrm{Sim}(z_i, z_{\text{ref}}) = (1 - \alpha)\,\mathrm{IoU}(b_i, b_{\text{ref}}) + \alpha\,\cos(s_i, s_{\text{ref}}),
\end{equation}
where $\alpha$ balances the two terms.
Queries with similarity above the pivot threshold are treated as close competitors and placed immediately after the pivot.
We then remove the pivot and its grouped competitors and repeat until all queries are ordered (Alg.~\ref{alg:cgqo} shown sequentially for clarity;  GPU-parallel details in Appendix~\ref{app:why_scale}).
This produces a stream where strong candidates appear early and their closest rivals are placed nearby. The gradients are stopped through ordering indices.

\vspace{-0.2cm}
\paragraph{Density-adaptive threshold.} To avoid using the same grouping threshold across different crowding levels, we assign a \textbf{pivot-specific} threshold:

{\setlength{\abovedisplayskip}{2pt}
 \setlength{\belowdisplayskip}{2pt}
 \setlength{\abovedisplayshortskip}{2pt}
 \setlength{\belowdisplayshortskip}{2pt}
\begin{equation}\label{eq:adaptive_thr}
T_i=\tau_{\min}+(\tau_{\max}-\tau_{\min})\cdot
\mathrm{norm}\!\left(\sum_{j\neq i}\mathbf{1}[\rho_{ij}\ge \tau_{\min}]\,\rho_{ij}\right),
\end{equation}
}

where $\rho_{ij}$ denotes the similarity in Eq.~\ref{eq:sim}, and $\tau_{\min}$ and $\tau_{\max}$ are the lower/upper bounds.
$\mathrm{norm}(\cdot)$ applies a fixed normalization and clamping to map the density score to $[0,1]$. Intuitively, denser neighborhoods yield larger $T_i$ (finer grouping), while sparser regions yield smaller $T_i$ to avoid over-splitting.
Unless specified otherwise, we use fixed hyperparameters for CGQO across all experiments. All hyperparameters and implementation details are provided in Appendix~\ref{app:cgqo_hypers}.

\setlength{\tabcolsep}{3pt}
\begin{table*}[!t]\footnotesize
\centering
\caption{Results on COCO \texttt{val2017} with ResNet-50 and 900 queries. Under unified $1\times$ / $2\times$ schedules, DETRNN consistently improves AP across baselines with comparable FLOPs / FPS.}
\begin{tabular}{l|ccccc|ccccccc} 
\toprule
Method & Backbone & Epochs & Params & FLOPs$\downarrow$ & FPS$\uparrow$
& AP & AP$_{50}$ & AP$_{75}$ & AP$_S$ & AP$_M$ & AP$_L$ \\
\midrule

\multirow{2}{*}{Def-DETR++~\cite{zhu2020deformable}} 
& \multirow{2}{*}{R50} 
& 12 & \multirow{2}{*}{47.5M} & \multirow{2}{*}{300G} & \multirow{2}{*}{21}
& 47.6 & 65.6 & 51.7 & 30.0 & 51.0 & 61.5 \\
&  & 24 &  &  &
& 49.3 & 67.9 & 53.7 & 32.0 & 52.6 & 63.7 \\
\addlinespace[4pt]

\multirow{2}{*}{Def-DETRNN++ (ours)} 
& \multirow{2}{*}{R50} 
& 12 & \multirow{2}{*}{47.6M} & \multirow{2}{*}{301G} & \multirow{2}{*}{20}
& \textbf{48.7} \textcolor{blue}{(+1.1)} & 67.0 & 53.2 & 32.3 & 52.4 & 63.2 \\
&  & 24 &  &  &
& \textbf{50.1} \textcolor{blue}{(+0.8)} & 68.7 & 54.8 & 33.7 & 53.2 & 64.7 \\
\midrule

\multirow{2}{*}{DINO~\cite{zhang2022dino}} 
& \multirow{2}{*}{R50} 
& 12 & \multirow{2}{*}{47.5M} & \multirow{2}{*}{300G} & \multirow{2}{*}{21}
& 49.0 & 66.6 & 53.5 & 32.0 & 52.3 & 63.0 \\
&  & 24 &  &  &
& 50.4 & 68.3 & 54.8 & 33.3 & 53.7 & 64.8 \\
\addlinespace[4pt]

\multirow{2}{*}{DINO-RNN (ours)} 
& \multirow{2}{*}{R50} 
& 12 & \multirow{2}{*}{47.6M} & \multirow{2}{*}{301G} & \multirow{2}{*}{20}
& \textbf{50.0} \textcolor{blue}{(+1.0)} & 68.1 & 54.0 & 32.9 & 53.2 & 64.3 \\
&  & 24 &  &  &
& \textbf{51.1} \textcolor{blue}{(+0.7)} & 69.0 & 55.9 & 34.3 & 54.2 & 65.2 \\

\midrule

\multirow{2}{*}{MS-DETR~\cite{Zhao2024MSDETRED}} 
& \multirow{2}{*}{R50} 
& 12 & \multirow{2}{*}{47.5M} & \multirow{2}{*}{309G} & \multirow{2}{*}{21}
& 50.0 & 67.3 & 54.4 & 31.6 & 53.2 & 64.0 \\
&  & 24 &  &  &
& 50.9 & 68.4 & 56.1 & 34.7 & 54.3 & 65.1 \\
\addlinespace[4pt]

\multirow{2}{*}{MS-DETRNN (ours)} 
& \multirow{2}{*}{R50} 
& 12 & \multirow{2}{*}{47.6M} & \multirow{2}{*}{310G} & \multirow{2}{*}{21}
& \textbf{50.5} \textcolor{blue}{(+0.5)} & 68.2 & 55.4 & 34.2 & 54.3 & 64.2 \\
&  & 24 &  &  &
& \textbf{51.4} \textcolor{blue}{(+0.5)} & 69.2 & 56.3 & 33.4 & 54.9 & 65.6 \\
\midrule

LoRA-DETR~\cite{zhang2026integrating} & R50 & 12 & 48.6M & 301G & 16 & 52.5 &  69.8 & 57.3 & 37.2 & 56.5 & 66.8 \\

LoRA-DETRNN (ours) & R50 & 12 & 48.7M & 301G & 19 & \textbf{53.0} \textcolor{blue}{(+0.5)} &  70.2 & 57.6 & 36.3 & 56.8 & 67.5 \\

\bottomrule
\end{tabular}
\label{tab:main_results_r50}
\end{table*}

\setlength{\tabcolsep}{5.7pt}
\begin{table}[!t]\footnotesize
\centering
\caption{
Results with Swin-L backbone.}

\begin{tabular}{l|ccc|cccccc}
\toprule
Method & Backbone & Epochs & Queries
& AP & AP$_{50}$ & AP$_{75}$ & AP$_S$ & AP$_M$ & AP$_L$ \\
\midrule
Def-DETR++~\cite{zhu2020deformable} & Swin-L & 12 & 300 &  54.5 & 74.0 &  59.3  & 37.0 & 58.6 & 71.0 \\
Def-DETRNN++ (ours) & Swin-L & 12 & 300 & \textbf{55.6} \textcolor{blue}{(+1.1)} & 75.1 & 60.7 & 38.5 & 59.8 & 71.4 \\

LoRA-DETR~\cite{zhang2026integrating} & Swin-L & 12 & 900 & 58.2 & 76.2 & 63.4 & 41.8 & 63.0 & 74.4 \\
LoRA-DETRNN (ours) & Swin-L & 12 & 900 & \textbf{58.5} \textcolor{blue}{(+0.3)} & 76.5 & 64.0 & 42.5 & 63.1 & 74.9 \\

\bottomrule
\end{tabular}
\label{tab:relation_swin}
\end{table}

\subsection{Sequential Query Refinement with Bi-RNN}
\label{subsec:birnn}
Given the competition-ordered query stream, we refine queries using an RNN-based interaction module.
Self-attention updates all queries in parallel and does not leverage an explicit order.
In contrast, an RNN naturally models an ordered stream, so it can directly use the CGQO order to pass suppression signals from strong candidates to their nearby competitors, reducing redundant hypotheses.

Formally, let $\mathbf{Z}=\{z_1,z_2,\cdots,z_N\}\in\mathbb{R}^{N\times d}$ denote the ordered queries produced by CGQO.
An RNN processes the sequence step by step:
\begin{equation}\label{eq:rnn}
[o_i, h_i] = \mathrm{RNN}(z_i, h_{i-1}),
\end{equation}
where $h_i$ is the hidden state and $o_i$ is the refined output at step $i$.
To strengthen information flow, we adopt a bi-directional design by also processing the reversed sequence and combining the two directions.
In practice, we use modern RNN-style blocks (e.g., Mamba~\cite{Gu2023MambaLS}) for efficient sequence modeling.
As shown in Fig.~\ref{fig:pipeline}, the Bi-RNN replaces the self-attention sub-layer in the decoder, while cross-attention and FFN remain unchanged, making the module plug-and-play for DETR-family detectors. Implementation details of our Bi-RNN instantiation are provided in Appendix~\ref{app:iof_mamba}.

\section{Experiments}

\subsection{Settings}
\paragraph{Datasets.} We conduct experiments on MS-COCO 2017~\cite{lin2014microsoft} and CrowdHuman~\cite{Shao2018CrowdHumanAB}.
For COCO, we train on \texttt{train2017} (117K images) and evaluate on \texttt{val2017} (5K images) using standard mAP.
CrowdHuman is a crowded human detection benchmark with heavy occlusions; we also report results on its validation (4.4K images).

\paragraph{Implementation details.}
Since implementations of recent detectors can differ substantially, we adapt four widely used baselines for fair comparison: Def-DETR++~\cite{zhu2020deformable}, MS-DETR~\cite{Zhao2024MSDETRED}, DINO~\cite{zhang2022dino}, and LoRA-DETR~\cite{zhang2026integrating}.
Unless otherwise specified, experiments use the 1$\times$ schedule of 12 epochs and follow the training recipe of Def-DETR++.
We use a total batch size of 16, learning rates of $2\times10^{-5}$ for the backbone and $2\times10^{-4}$ for the Transformer encoder--decoder, and a weight decay of $2\times10^{-5}$.
For the sequential refinement module, we instantiate the Bi-RNN with Mamba~\cite{Gu2023MambaLS} in most experiments.
Module configurations, hyperparameter settings, and sensitivity results are provided in Appendix~\ref{app:v_rnn}--\ref{app:efficiency_metrics}.

\subsection{Main Results}

We evaluate our plug-and-play module on 4 representative DETR variants.
As shown in Tabs.~\ref{tab:main_results_r50} and~\ref{tab:relation_swin}, DETRNN consistently improves AP across all baselines with comparable efficiency.
These results show that explicit query-competition modeling is broadly beneficial and transferable, with full-recipe \textbf{SOTA} comparison in Appendix~\ref{app:full_recipe_context}.
A detailed analysis of each baseline is provided in the following paragraphs.

\noindent \textbf{Def-DETR++.} Our plug-and-play module consistently improves Def-DETR++ under the same training schedule. With 900 queries and a $1\times$ schedule, it increases AP from 47.6 to 48.7 (+1.1). Under the $2\times$ schedule, it improves the baseline from 49.3 to 50.1 (+0.8), indicating the gain persists when the baseline is further strengthened.

\noindent \textbf{DINO.} We observe consistent gains when integrating our module into this baseline. With a $1\times$ schedule, AP increases from 49.0 to 50.0 (+1.0). With a $2\times$ schedule, AP improves from 50.4 to 51.1 (+0.7), showing that our method remains effective across training schedules.

\noindent \textbf{MS-DETR.} Despite being a strong baseline, MS-DETR also benefits from our module. At $1\times$, AP increases from 50.0 to 50.5 (+0.5). At $2\times$, it improves from 50.9 to 51.4 (+0.5), demonstrating stable gains even with stronger training.

\noindent \textbf{LoRA-DETR.} DETRNN is further evaluated on LoRA-DETR~\cite{zhang2026integrating}, a new SOTA DETR baseline. Using its best variant built on Relation-DETR~\cite{hou2024relation}, under the $1\times$ schedule, DETRNN improves AP from 52.5 to 53.0 (+0.5). Moreover, by replacing baseline's more complex query interaction with our refinement, we \textbf{improve} efficiency with unchanged FLOPs and higher FPS (16$\rightarrow$19).

\noindent \textbf{Results with Swin-L backbone.}
Tab.~\ref{tab:relation_swin} reports the performance of our variants under the Swin-L~\cite{liu2021SwinTH} backbone.
Def-DETRNN++ achieves a clear gain over Def-DETR++ with 300 queries, improving AP from 54.5 to 55.6 (+1.1), with consistent improvements on AP$_{50}$ and AP$_{75}$.
Moreover, LoRA-DETRNN also yields a consistent boost over LoRA-DETR under the 900-query setting, reaching 58.5 AP (+0.3).
These results suggest that the proposed refinement generalizes well to stronger backbones and remains compatible with different detector designs.

\begin{table}[t]
\footnotesize
\centering
\setlength{\tabcolsep}{2.2pt}
\caption{\textbf{1}-decoder de-duplication stress test (900 queries, $1\times$ schedule).
We report AP and +NMS AP (AP after NMS@0.8); $\Delta$AP denotes the NMS gain, where a larger value indicates more residual duplicates.
DETRNN improves AP while sharply reducing $\Delta$AP, suggesting stronger decoder-internal de-duplication.
}
\label{tab:one_decoder_nms}
\begin{minipage}[t]{0.32\linewidth}
\centering
\begin{tabular}{c|ccc}
\toprule
\multicolumn{4}{c}{Def-DETR++ (1-decoder)} \\
\cmidrule(lr){1-4}
 & AP & +NMS & $\Delta$AP$\downarrow$ \\
\midrule
Base  & 40.6 & 43.0 & 2.4 \\
+Ours & \textbf{46.3} \textcolor{blue}{(+5.7)} & 46.6 & \textbf{0.3} \textcolor{blue}{(-2.1)}\\
\bottomrule
\end{tabular}
\end{minipage}\hfill
\begin{minipage}[t]{0.32\linewidth}
\centering
\begin{tabular}{c|ccc}
\toprule
\multicolumn{4}{c}{DINO (1-decoder)} \\
\cmidrule(lr){1-4}
 & AP & +NMS & $\Delta$AP$\downarrow$ \\
\midrule
Base  & 42.2 & 44.7 & 2.5 \\
+Ours & \textbf{46.8} \textcolor{blue}{(+4.6)} & 47.0  & \textbf{0.2} \textcolor{blue}{(-2.3)} \\
\bottomrule
\end{tabular}
\end{minipage}\hfill
\begin{minipage}[t]{0.32\linewidth}
\centering
\begin{tabular}{c|ccc}
\toprule
\multicolumn{4}{c}{MS-DETR (1-decoder)} \\
\cmidrule(lr){1-4}
 & AP & +NMS & $\Delta$AP$\downarrow$ \\
\midrule
Base  & 40.1 & 43.1 & 3.0 \\
+Ours & \textbf{43.8} \textcolor{blue}{(+3.7)} & 44.7 & \textbf{0.9} \textcolor{blue}{(-2.1)} \\
\bottomrule
\end{tabular}
\end{minipage}
\end{table}

\begin{figure}[t]
\centering
\begin{subfigure}[t]{0.31\textwidth}
  \centering
  \includegraphics[width=\linewidth,height=2.7cm,keepaspectratio]{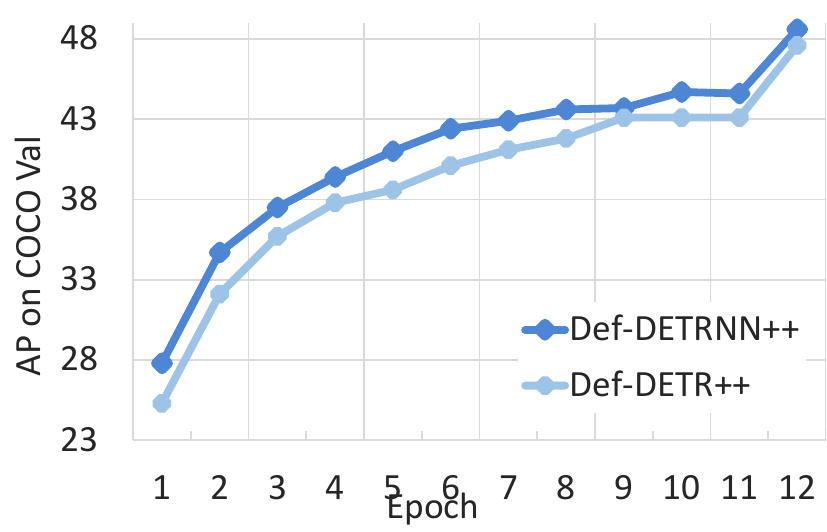}
  \phantomsubcaption\label{fig:train_diag_ap}
  \subcaption*{(\subref{fig:train_diag_ap})}
\end{subfigure}\hfill
\begin{subfigure}[t]{0.31\textwidth}
  \centering
  \includegraphics[width=\linewidth,height=2.7cm,keepaspectratio]{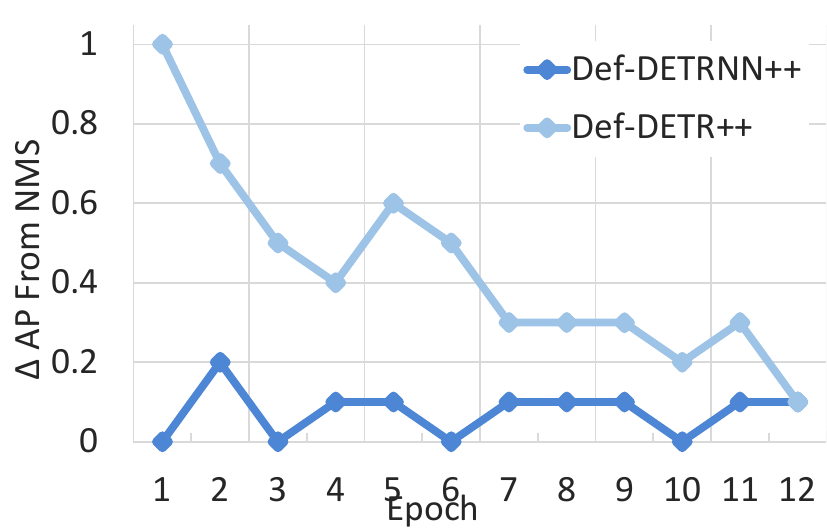}
  \phantomsubcaption\label{fig:train_diag_nms}
  \subcaption*{(\subref{fig:train_diag_nms})}
\end{subfigure}\hfill
\begin{subfigure}[t]{0.31\textwidth}
  \centering
  \includegraphics[width=\linewidth,height=2.7cm,keepaspectratio]{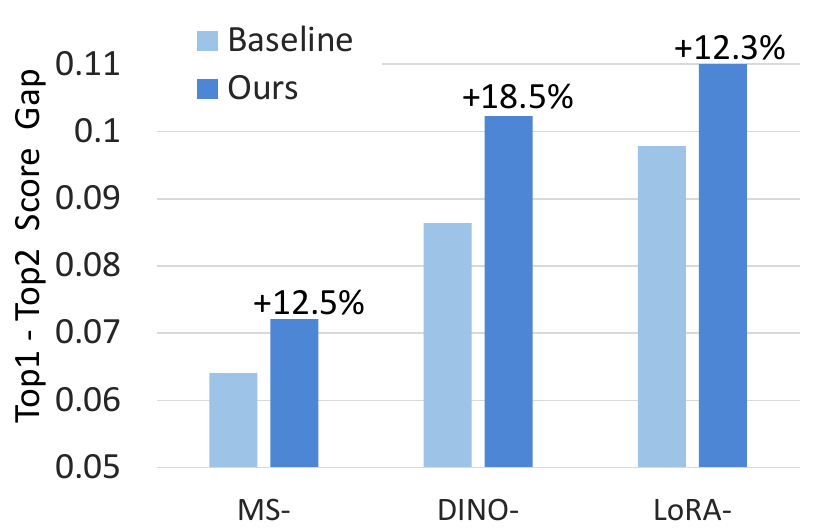}
  \phantomsubcaption\label{fig:train_diag_gap}
  \subcaption*{(\subref{fig:train_diag_gap})}
\end{subfigure}
    \caption{Training dynamics and competition indicators on COCO \texttt{val} (\textbf{6}-decoder).
(a) Validation AP across epochs for Def-DETR++ and ours.
(b) AP gain from evaluation-time NMS@0.8; larger values indicate more residual duplicates.
(c) Top1--Top2 score gap, where Top1/Top2 are the best and second-best predictions matched to the same GT; a larger gap indicates clearer separation between the correct detection and its closest competitor.}
    \label{fig:train_diag}
\vspace{-0.1cm}
\end{figure}

\subsection{De-duplication and Training Dynamics}

We verify decoder-internal de-duplication beyond final AP using two complementary checks: a shallow-decoder probe and training-time diagnostics.

\noindent\textbf{Shallow-decoder probe.}
Tab.~\ref{tab:one_decoder_nms} reports a \textbf{1-decoder} stress test (see Appendix~\ref{app:1-decoder} for settings), where redundancy is strongest because DETR detectors typically rely on multiple decoders to gradually suppress duplicates.
We build CGQO from encoder proposals and evaluate the 1st decoder output.
DETRNN brings large AP gains (+5.7/+4.6/+3.7) and even outperforms the baseline with NMS.
Since NMS mainly removes duplicate predictions, a larger $\Delta$AP indicates more residual redundancy in the outputs.
Baselines show large $\Delta$AP (2.4--3.0), while DETRNN reduces it to 0.2--0.9, indicating that far fewer duplicate predictions remain for NMS to remove.

\noindent\textbf{Training dynamics.}
In the standard \textbf{6-decoder} setting, Fig.~\ref{fig:train_diag} tracks how de-duplication evolves during training.
DETRNN achieves higher AP from early epochs (Fig.~\ref{fig:train_diag}(a)) and much smaller $\Delta$AP from NMS@0.8, especially early in training (Fig.~\ref{fig:train_diag}(b)).
This indicates stable optimization and earlier suppression of redundant competitors.
DETRNN also enlarges the Top1--Top2 score gap (Fig.~\ref{fig:train_diag}(c)), showing clearer separation between the selected prediction and its closest competitor, which stabilizes one-to-one matching.

\begin{table}[t]
\footnotesize
\centering
\setlength{\tabcolsep}{4.0pt}
\caption{
Ablations on baselines MS- and DINO-. BR denotes replacing self-attention with Bi-RNN, and CGQO denotes Competition-Guided Query Ordering.
Using BR or CGQO alone yields negligible gains, while combining them leads to clear improvements, confirming that Bi-RNN can effectively model inter-query dependencies under an explicit competition order.}
\label{tab:ablation_ms_detr_dino}
\begin{minipage}[t]{0.45\linewidth}
\centering
\begin{tabular}{c|cc|ccc}
\toprule
\multicolumn{6}{c}{DINO-} \\
\midrule
Method & BR & CGQO & AP & AP$_{50}$ & AP$_{75}$ \\
\midrule
1 &  &  & 49.0 & 66.6 & 53.5 \\
2 & \checkmark &  & 49.2 \textcolor{blue}{(+0.2)} & 66.9 & 53.8 \\
3 &  & \checkmark & 49.0 & 66.6 & 53.5 \\
4 & \checkmark & \checkmark & \textbf{50.0} \textcolor{blue}{(+1.0)} & \textbf{68.1} & \textbf{54.0} \\
\bottomrule
\end{tabular}
\end{minipage}
\hfill
\begin{minipage}[t]{0.45\linewidth}
\centering
\begin{tabular}{c|cc|ccc}
\toprule
\multicolumn{6}{c}{MS-} \\
\midrule
Method & BR & CGQO & AP & AP$_{50}$ & AP$_{75}$ \\
\midrule
1 &  &  & 50.0 & 67.3 & 54.4 \\
2 & \checkmark &  & 50.0 & 67.3 & 53.3 \\
3 &  & \checkmark & 50.0 & 67.3 & 54.4 \\
4 & \checkmark & \checkmark & \textbf{50.5} \textcolor{blue}{(+0.5)} & \textbf{68.2} & \textbf{55.4} \\
\bottomrule
\end{tabular}
\end{minipage}
\end{table}

\begin{table}[t]
\vspace{-0.1cm}
\centering
\footnotesize
\setlength{\tabcolsep}{16pt}
\caption{Ordering variants for Bi-RNN refinement measured by \textbf{AP}.
Unordered: Method~2 in Tab.~\ref{tab:ablation_ms_detr_dino}.
Conf-rank: confidence-only ranking.
Group-IoU: pivot-and-group with IoU-only similarity. CGQO: Method~4 in Tab.~\ref{tab:ablation_ms_detr_dino}. Parentheses report AP changes relative to the Unordered variant.
}
\label{tab:ordering_variants}
\begin{tabular}{l | ccccc}
\toprule
Detector & Unordered & Random & Conf-rank & Group-IoU & CGQO \\
\midrule
DINO- & 49.2 & 48.8 (\textcolor{red}{-0.4}) & 49.3 (\textcolor{blue}{+0.1}) & 49.8 (\textcolor{blue}{+0.6}) & \textbf{50.0} (\textcolor{blue}{+0.8})  \\
MS-   & 50.0 & 49.8 (\textcolor{red}{-0.2}) & 50.0 & 50.2 (\textcolor{blue}{+0.2}) & \textbf{50.5} (\textcolor{blue}{+0.5}) \\
\bottomrule
\end{tabular}
\end{table}
\vspace{-0.1cm}

\begin{figure}[b]
\centering
\begin{subfigure}[t]{0.53\linewidth}
  \vspace{0pt} 
  \centering
  \footnotesize
  \setlength{\tabcolsep}{5pt}
  \renewcommand{\arraystretch}{0.92}
  \begin{tabular}{l|c|ccc}
    \toprule
    Detector & Backbone & AP & AP$_{50}$ & AP$_{75}$ \\
    \midrule
    Attn (Base) & R50 & 47.6 & 65.6 & 51.7 \\
    \midrule
    Mamba & R50 & 48.7 (\textcolor{blue}{+1.1}) & 67.0 & 53.2 \\
    RWKV  & R50 & 48.6 (\textcolor{blue}{+1.0}) & 66.8 & 53.1 \\
    xLSTM & R50 & 48.4 (\textcolor{blue}{+0.8}) & 66.6 & 52.8 \\
    TTT   & R50 & 48.7 (\textcolor{blue}{+1.1}) & 67.3 & 52.8 \\
    \bottomrule
  \end{tabular}
  \vspace{0.16cm}
  \subcaption{AP with different sequential modules}
  \label{fig:rnn_variants_ap}
\end{subfigure}\hfill
\begin{subfigure}[t]{0.44\linewidth}
  \vspace{0pt} 
  \centering
  \hspace{0.06 cm}
  \includegraphics[width=\linewidth,height=2.6cm,keepaspectratio]{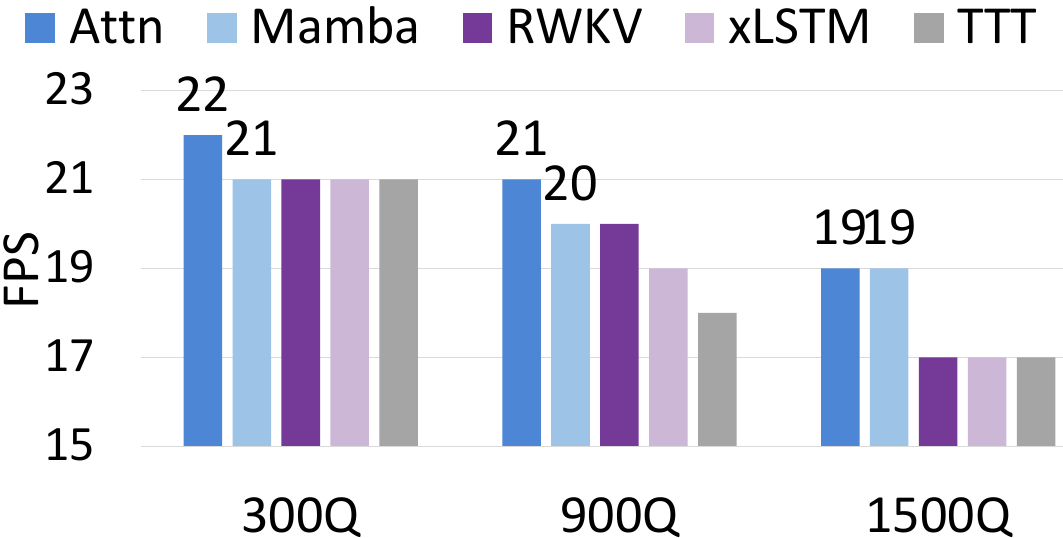}
  \subcaption{FPS vs.\ number of queries (300/900/1500)}
  \label{fig:rnn_variants_fps}
\end{subfigure}
\caption{RNN variants and efficiency scaling on Def-DETRNN++ detector ($1\times$ schedule).
(a) Different RNN-style modules yield stable gains over baseline.
(b) The Mamba variant has controllable FPS overhead, with the gap to Attn narrowing as the number of queries increases.}
\label{fig:rnn_variants}
\end{figure}

\subsection{Ablation Study}

\noindent\textbf{Component ablation.}
Tab.~\ref{tab:ablation_ms_detr_dino} ablates Bi-RNN (BR) and Competition-Guided Query Ordering (CGQO).
Bi-RNN alone (Method~2) brings limited benefit because, without an informative order, sequential refinement has little structure to exploit.
CGQO alone (Method~3) is also ineffective, since self-attention is permutation-equivariant and processes queries in parallel.
In contrast, combining them (Method~4) yields clear gains (e.g., +1.0 on DINO- and +0.5 on MS-), showing that the improvement comes from the interaction between explicit competition ordering and sequential refinement, rather than from either component alone.

\noindent\textbf{Ordering variants.}
To isolate the role of ordering itself, Tab.~\ref{tab:ordering_variants} keeps Bi-RNN fixed and varies only the input order.
Unordered and CGQO correspond to Methods~2 and~4 in Tab.~\ref{tab:ablation_ms_detr_dino}, respectively.
Conf-rank keeps only the confidence ranking step in line~3 of Alg.~\ref{alg:cgqo}, and Group-IoU keeps the pivot-and-group procedure but uses IoU-only similarity in Eq.~\ref{eq:sim} ($\alpha{=}0$).
With unsuitable orders, sequential refinement helps little or even hurts: Random ordering degrades AP (-0.4/-0.2), and Conf-rank yields only marginal gains (+0.1/+0.0).
A more competition-aware order improves results (Group-IoU), while \textbf{CGQO} performs best.
This rules out a simple capacity explanation: the same Bi-RNN becomes effective only when close competitors are placed nearby in a meaningful order.

\noindent\textbf{RNN variants and query scaling.}
Fig.~\ref{fig:rnn_variants} studies whether the gain depends on a specific sequential module.
With CGQO fixed, Mamba, RWKV~\cite{peng2023RWKVRR}, xLSTM~\cite{beck2024xLSTMEL}, and TTT~\cite{sun2024LearningT} all perform stably (48.4--48.7 AP), consistently outperforming the Attn baseline (47.6 AP).
This suggests that the key factor is not a particular RNN design, but the combination of competition-aware ordering and sequential refinement.
Fig.~\ref{fig:rnn_variants}(b) reports FPS when scaling the number of queries from 300 to 1500.

As $N$ increases, the Mamba variant maintains comparable FPS to Attn in the detector, and Appendix~\ref{app:profiling} shows better memory scaling and competitive module runtime under larger query counts.

\begin{table}[t]
\footnotesize
\centering
\setlength{\tabcolsep}{14pt}
\caption{Threshold sensitivity on CrowdHuman dataset. Each entry reports \textbf{AP} under a specific setting. NMS is sensitive to the IoU threshold: a lower $\tau$ (0.7) can reduce AP, while higher $\tau$ brings only marginal gains. In contrast, our method consistently improves accuracy without threshold tuning.}
\label{tab:crowdhuman_nms}
\begin{tabular}{l|ccccc}
\toprule
Detector & w/o NMS & +NMS@0.7 & +NMS@0.8 & +NMS@0.9 & +Ours \\
\midrule
Def-DETR++ & 46.5 & 46.2~(\textcolor{red}{-0.3}) & 46.7~(\textcolor{blue}{+0.2}) & 46.6~(\textcolor{blue}{+0.1}) & \textbf{47.6}~(\textcolor{blue}{+1.1}) \\
MS-DETR    & 48.4 & 48.3~(\textcolor{red}{-0.1}) & 48.6~(\textcolor{blue}{+0.2}) & 48.6~(\textcolor{blue}{+0.2}) & \textbf{49.2}~(\textcolor{blue}{+0.8}) \\
\bottomrule
\end{tabular}
\end{table}

\begin{table}[t]
\centering
\footnotesize
\setlength{\tabcolsep}{14pt}
\caption{COCO consistency check (AP). NMS brings only marginal gains (+0.1--0.2) for the baselines.}
\label{tab:coco_nms}
\begin{tabular}{l|ccccc}
\toprule
Detector & w/o NMS & +NMS@0.7 & +NMS@0.8 & +NMS@0.9 & +Ours \\
\midrule
Def-DETR++  & 47.6 & 47.6 & 47.7 (\textcolor{blue}{+0.1}) & 47.7 (\textcolor{blue}{+0.1}) & \textbf{48.7} (\textcolor{blue}{+1.1}) \\
MS-DETR & 50.0 & 50.1 (\textcolor{blue}{+0.1}) & 50.2 (\textcolor{blue}{+0.2}) & 50.2 (\textcolor{blue}{+0.2}) & \textbf{50.5} (\textcolor{blue}{+0.5}) \\
\bottomrule
\end{tabular}
\end{table}

\subsection{Crowded-Scene Robustness}
Tab.~\ref{tab:crowdhuman_nms} compares our method with applying NMS under the $1\times$ schedule on CrowdHuman, a crowded benchmark with heavy box overlaps, where NMS becomes sensitive to the IoU threshold.
With a lower threshold ($\tau{=}0.7$), NMS is more aggressive and can remove true positives, leading to AP drops of $0.1\!\sim\!0.3$ across baselines.
With a higher threshold ($\tau{=}0.8/0.9$), NMS is less destructive but leaves more duplicates, so the gains are limited (typically $+0.1\!\sim\!+0.2$).
In contrast, our method suppresses redundant competitors inside the decoder and yields consistent improvements of $+0.8\!\sim\!+1.1$ AP without NMS or threshold tuning.

\paragraph{Consistency on COCO.}
Following Tab.~\ref{tab:crowdhuman_nms}, Tab.~\ref{tab:coco_nms} repeats the same NMS-threshold check on COCO.
Like CrowdHuman, NMS on COCO provides only marginal gains for the baselines (+0.1--0.2 AP).
Our method improves AP by +1.1/+0.5 without relying on post-processing threshold tuning.

\subsection{Replacement on Competition-Aware Decoders}
Tab.~\ref{tab:compatibility_competition_aware} evaluates DETRNN on three representative competition-aware decoders under the $2\times$ schedule: pairwise-interaction methods EASE-~\cite{gao2024EASEDETRET} and Relation-~\cite{hou2024relation}, and the query-selection method DDQ~\cite{Zhang_2023_CVPR}.
Because these baselines differ beyond query interaction, we use a replacement setting: \emph{Base} is the original full model, while \emph{+Ours} keeps its training recipe and replaces only the competition-specific decoder operation.
DETRNN improves AP on all three baselines (+0.3/+0.5/+0.4) and also increases throughput (18$\rightarrow$19, 16$\rightarrow$19, and 15$\rightarrow$17 FPS) with comparable FLOPs.
These results suggest that ordered sequential refinement can serve as an efficient alternative to existing competition-specific designs.
More details are provided in Appendix~\ref{app:compet_aware}.

\begin{table}[!t]
\footnotesize
\centering
\setlength{\tabcolsep}{2.3pt}
\caption{
Replacement on competition-aware decoders.
DETRNN improves pairwise-interaction methods (EASE-/Relation-) and a query-selection method (DDQ) with higher throughput.}

\label{tab:compatibility_competition_aware}
\begin{minipage}[t]{0.3\linewidth}
\centering
\begin{tabular}{c|ccc}
\toprule
\multicolumn{4}{c}{EASE-DETR} \\
\cmidrule(lr){1-4}
 & FLOPs$\downarrow$ & FPS$\uparrow$ & AP \\
\midrule
Base  & 301G & 18 & 49.6 \\
+Ours & 301G & \textbf{19} & \textbf{49.9} \textcolor{blue}{(+0.3)} \\
\bottomrule
\end{tabular}
\end{minipage}%
\hfill
\begin{minipage}[t]{0.3\linewidth}
\centering
\begin{tabular}{c|ccc}
\toprule
\multicolumn{4}{c}{Relation-DETR} \\
\cmidrule(lr){1-4}
 & FLOPs$\downarrow$ & FPS$\uparrow$ & AP \\
\midrule
Base  & 301G & 16 & 52.1 \\
+Ours & 300G & \textbf{19} & \textbf{52.6} \textcolor{blue}{(+0.5)} \\
\bottomrule
\end{tabular}
\end{minipage}%
\hfill
\begin{minipage}[t]{0.3\linewidth}
\centering
\begin{tabular}{c|ccc}
\toprule
\multicolumn{4}{c}{DDQ} \\
\cmidrule(lr){1-4}
 & FLOPs$\downarrow$ & FPS$\uparrow$ & AP \\
\midrule
Base  & 300G & 15 & 52.0 \\
+Ours & 300G & \textbf{17} & \textbf{52.4} \textcolor{blue}{(+0.4)} \\
\bottomrule
\end{tabular}
\end{minipage}
\end{table}

\subsection{Stage Placement}
\label{app:stage_place}

Tab.~\ref{tab:stage_placement_all} studies where to insert our plug-in module
across decoder stages.
Def-DETR++ and DINO perform best with the last two stages, whereas MS-DETR and
LoRA-DETR favor only the final stage.
Restricting the module to these late stages also avoids unnecessary recurrent
computation and keeps the additional refinement lightweight, with higher FPS
than applying it to all decoder stages.
This is consistent with the stage-wise AP/AR100 analysis in
Appendix~\ref{app:stage_analysis}, where recall largely saturates before the
final stage while AP continues to improve.
We therefore use last-two-stage insertion for Def-DETRNN++ and DINO-RNN, and
last-stage insertion for MS-DETRNN and LoRA-DETRNN.
CGQO is recomputed from the current predictions before each inserted stage.

\begin{table}[t]
\footnotesize
\centering
\caption{
Stage placement study across different baselines.
``Stage'' indicates the decoder stages where our module is inserted.
}
\label{tab:stage_placement_all}

\begin{minipage}{1\linewidth}
\centering

\begin{minipage}[t]{0.2\linewidth}
\centering
\begin{tabular*}{\linewidth}{@{\extracolsep{\fill}}c|cc@{}}
\toprule
\multicolumn{3}{c}{Def-DETRNN++} \\
\cmidrule(lr){1-3}
Stage & FPS$\uparrow$ & AP \\
\midrule
All    & 18 & 48.4 \\
Last 2 & 20 & \textbf{48.7} \\
Last 1 & \textbf{21} & 48.4 \\
\bottomrule
\end{tabular*}
\end{minipage}
\hfill
\begin{minipage}[t]{0.2\linewidth}
\centering
\begin{tabular*}{\linewidth}{@{\extracolsep{\fill}}c|cc@{}}
\toprule
\multicolumn{3}{c}{DINO-RNN} \\
\cmidrule(lr){1-3}
Stage & FPS$\uparrow$ & AP \\
\midrule
All    & 18 & 49.3 \\
Last 2 & 20 & \textbf{50.0} \\
Last 1 & \textbf{21} & 49.7 \\
\bottomrule
\end{tabular*}
\end{minipage}
\hfill
\begin{minipage}[t]{0.2\linewidth}
\centering
\begin{tabular*}{\linewidth}{@{\extracolsep{\fill}}c|cc@{}}
\toprule
\multicolumn{3}{c}{MS-DETRNN} \\
\cmidrule(lr){1-3}
Stage & FPS$\uparrow$ & AP \\
\midrule
All    & 18 & 49.7 \\
Last 2 & 20 & 50.2 \\
Last 1 & \textbf{21} & \textbf{50.5} \\
\bottomrule
\end{tabular*}
\end{minipage}
\hfill
\begin{minipage}[t]{0.2\linewidth}
\centering
\begin{tabular*}{\linewidth}{@{\extracolsep{\fill}}c|cc@{}}
\toprule
\multicolumn{3}{c}{LoRA-DETRNN} \\
\cmidrule(lr){1-3}
Stage & FPS$\uparrow$ & AP \\
\midrule
All    & 16 & 52.7 \\
Last 2 & 18 & 52.9 \\
Last 1 & \textbf{19} & \textbf{53.0} \\
\bottomrule
\end{tabular*}
\end{minipage}

\end{minipage}
\end{table}

\begin{figure}[t]
\centering
\vspace{-0.1cm}
\begin{subfigure}[t]{0.34\linewidth}
  \vspace{0pt}
  \centering
  \includegraphics[width=\linewidth]{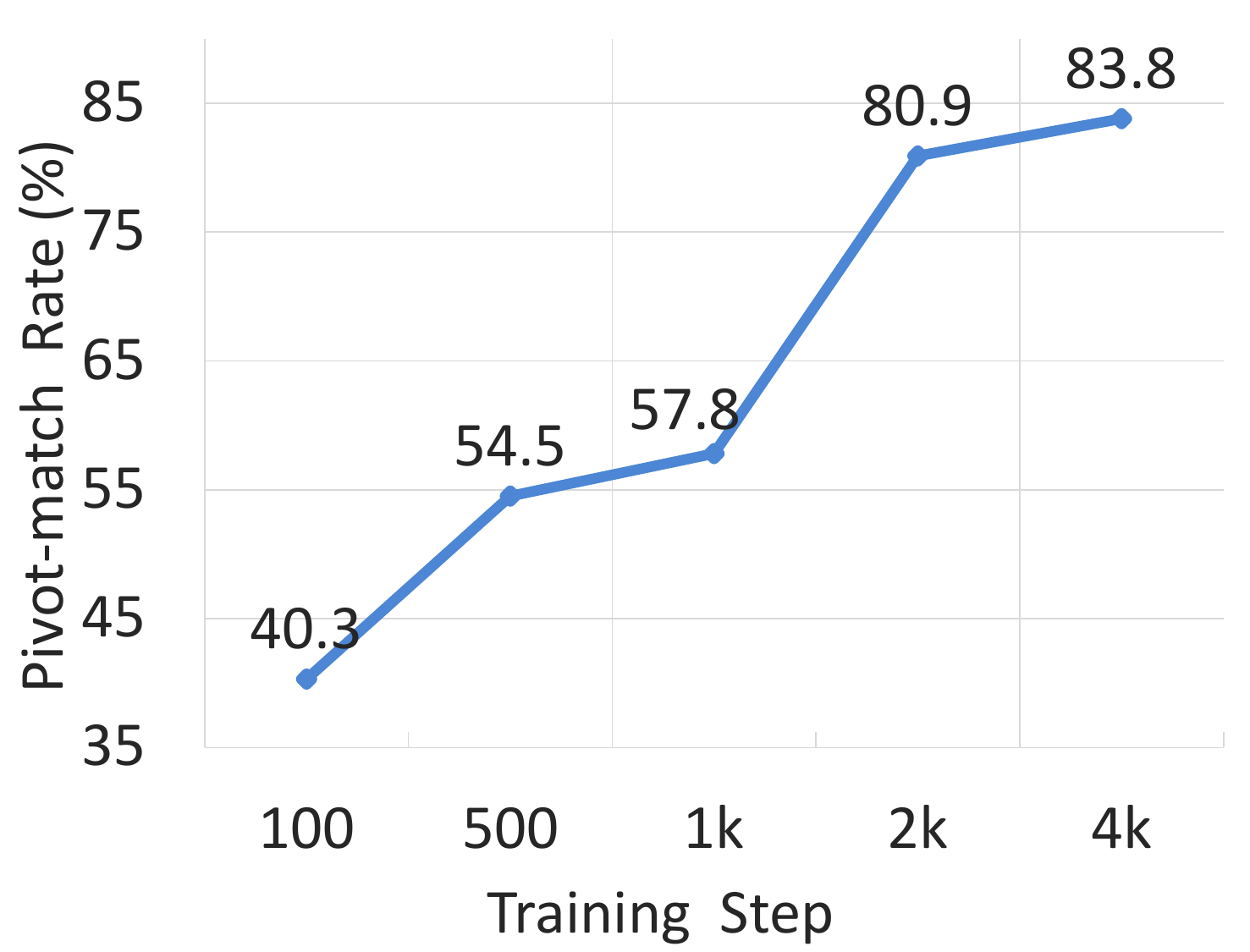}
  \subcaption*{(a)}
\end{subfigure}
\hfill
\begin{subfigure}[t]{0.26\linewidth}
\vspace{0.18cm}
  \centering
  \includegraphics[width=\linewidth]{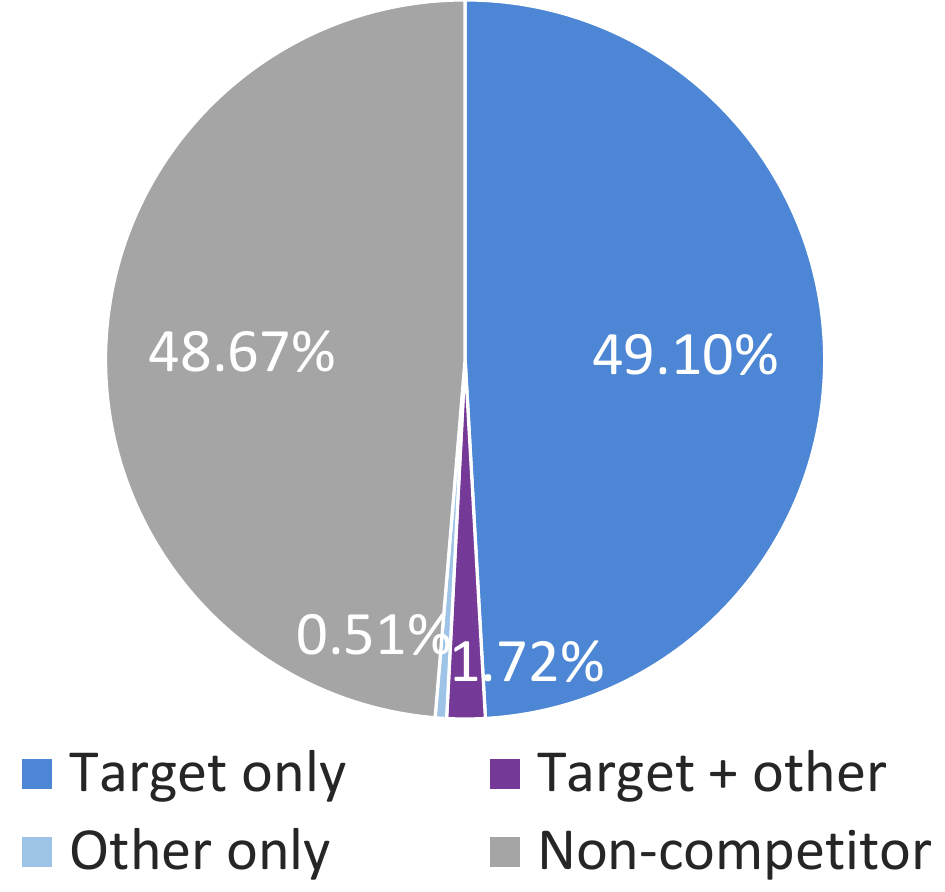}
  \subcaption*{(b)}
\end{subfigure}
\hfill
\begin{subfigure}[t]{0.31\linewidth}
\vspace{0.4cm}
  \centering
  \footnotesize
  \setlength{\tabcolsep}{11pt}
  \renewcommand{\arraystretch}{1.7}

\begin{tabular}{@{}l| r@{}}
  \toprule
  Metric & Value \\
  \midrule
  Pivot-match rate      & 89.24\% \\
  Same-GT coverage       & 78.85\% \\
  Group vs.\ GT count ($r$) & 0.863 \\
  \bottomrule
\end{tabular}
\vspace{0.49cm}
  \subcaption*{(c)}
\end{subfigure}

\caption{
Mechanistic analysis of CGQO grouping.
(a) Pivot-match evolution during the first training epoch.
(b) Composition of the Hungarian-matched query's CGQO group.
(c) Summary statistics of object-centered grouping.
}
\label{fig:cgqo_group_analysis}
\vspace{-0.1cm}
\end{figure}

\subsection{Mechanistic Analysis of CGQO Grouping}
\vspace{-0.15cm}

To examine whether CGQO forms object-centered competition groups, we analyze
their evolution, composition, and converged structure on the first 1,600 COCO
\texttt{val2017} images using Def-DETRNN++ ($1\times$).
Fig.~\ref{fig:cgqo_group_analysis}(a) uses first-epoch checkpoints, while
Figs.~\ref{fig:cgqo_group_analysis}(b,c) use the converged model.

\noindent\textbf{Early grouping evolution.}
We measure the pivot-match rate, the percentage of GT objects whose
Hungarian-matched query is also the pivot of its CGQO group; a higher rate
indicates more reliable pivot selection.
As shown in Fig.~\ref{fig:cgqo_group_analysis}(a), it rises from 40.3\% at
step 100 to 80.9\% at 2K and 83.8\% at 4K, showing rapid improvement during
early training; its converged value is reported in
Fig.~\ref{fig:cgqo_group_analysis}(c).

\noindent\textbf{Group composition.}
For each GT, we define a \emph{foreground competitor} as a query with box IoU
$\geq 0.5$ and GT-class score $\geq 0.05$, and examine the CGQO group containing
its Hungarian-matched query.
Among all members of this group, 49.10\% are competitors only for the target GT,
1.72\% for both the target and other GTs, and only 0.51\% exclusively for other
GTs; the remaining 48.67\% are non-competitor queries, as expected since CGQO
retains and groups all queries (Fig.~\ref{fig:cgqo_group_analysis}(b)).
Among foreground competitors, \textbf{99.01\%} are associated with the target GT, showing very little cross-object mixing.

\noindent\textbf{Object-centered grouping.}
The Hungarian-matched query is the group pivot in 89.24\% of cases, indicating
reliable pivot selection.
The corresponding CGQO group covers 78.85\% of the foreground competitors for
the same GT, showing that most same-object competitors are grouped together.
Across images, the number of groups containing foreground competitors strongly
correlates with the number of GT objects (Pearson $r=0.863$).
These complementary statistics support the object-centered structure of CGQO
grouping (Fig.~\ref{fig:cgqo_group_analysis}(c)).

\section{Conclusion}
We introduced DETRNN, which turns de-duplication in end-to-end detection into a learnable sequential refinement process.
By constructing a competition-aware query order and refining queries along it with an RNN module, DETRNN suppresses redundant predictions while keeping all predictions end-to-end.
To our knowledge, this is among the first attempts to treat de-duplication in set-based detection as a sequential refinement problem inside the DETR decoder.
Our results across diverse detectors validate a simple message: order matters---a well-structured competition order turns an unordered set into a meaningful sequence, making sequential models a natural fit for resolving duplicates in end-to-end prediction. A brief discussion of limitations is provided in Appendix~\ref{app:limitations}.

\newpage
\begin{ack}
This work was supported by the Science and Technology Commission of Shanghai Municipality under Grant
No.22511105800 and 22DZ2229004, the AI
project from the economic and information commission of
shanghai (Grant No. 2024-GZL-RGZN-01038), and ECNU
Multifunctional Platform for Innovation (001).

The authors declare no competing interests.
\end{ack}

\bibliographystyle{unsrt}
\bibliography{neurips_2026}

\newpage
\appendix

\section*{\centering Technical Appendices and Supplementary Material}

\begin{figure}[H]
    \centering
    \includegraphics[width=14cm]{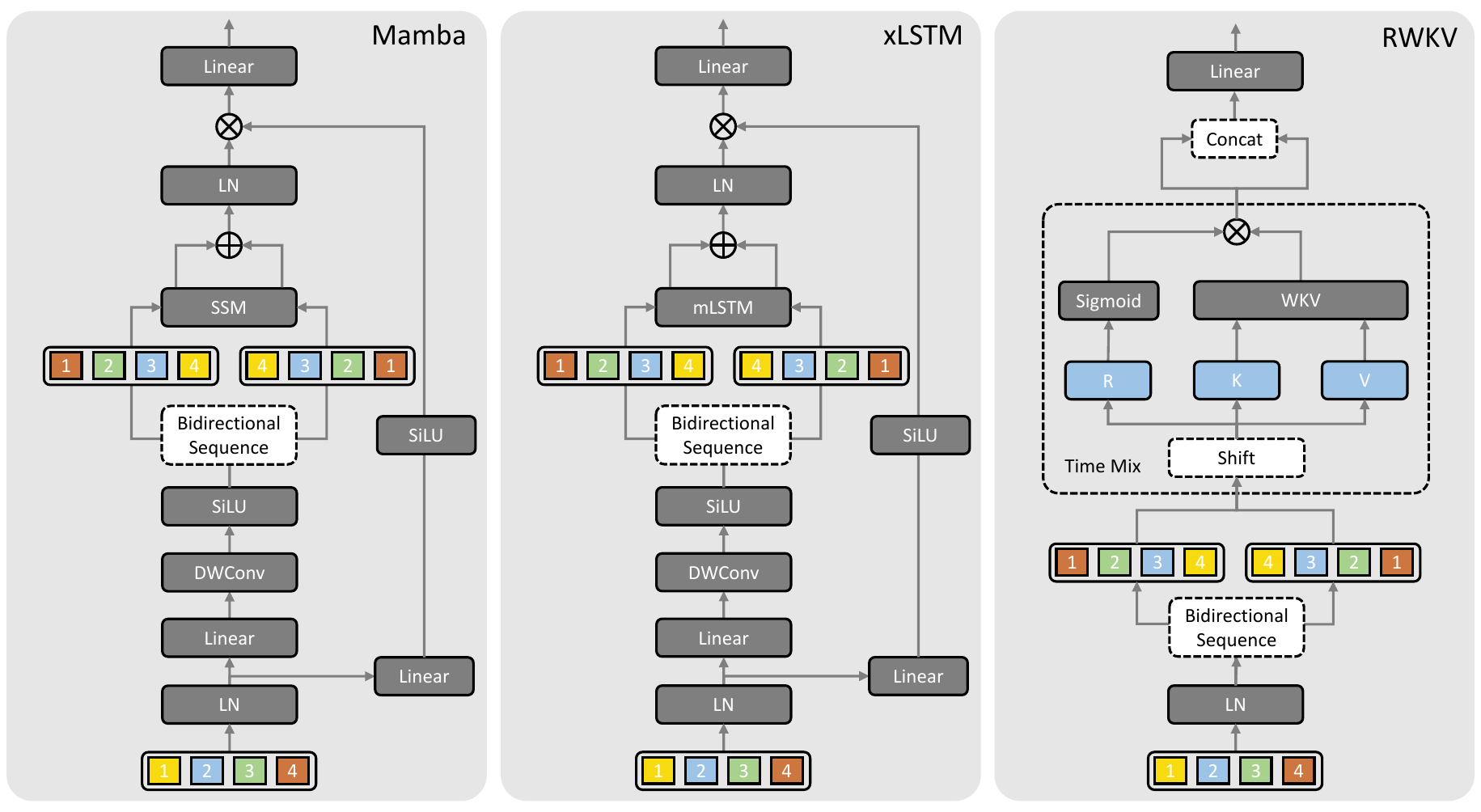}
    \caption{Structures of the three representative RNN variants used in our Bi-RNN module. TTT is also evaluated but omitted from the architecture illustration for space.
    }
    \label{fig:RNN}
\end{figure}

\section{Various RNN architectures}
\label{app:v_rnn}
Figure \ref{fig:RNN} illustrates the three RNN architectures used in our method, each differing in their query interaction mechanisms but all incorporating bidirectional processing to enhance query interactions. Mamba ~\cite{Gu2023MambaLS} maintains its original structure with bidirectional query processing, utilizing components like SiLU activation, depthwise convolution, and layer normalization. xLSTM \cite{beck2024xLSTMEL} extends LSTM with exponential gating and matrix-memory mechanisms such as mLSTM. Similarly, RWKV \cite{peng2023RWKVRR} incorporates bidirectional processing with weighted key-value mixing and temporal shifts. By integrating these bidirectional mechanisms, xLSTM and RWKV further enhance the model's ability to capture relationships within the object query sequence.

\section{CGQO Hyperparameters}
\label{app:cgqo_hypers}

\begin{figure}[t]
\centering

\begin{subfigure}[t]{0.48\linewidth}
  \centering
  \includegraphics[width=\linewidth,height=2.6cm,keepaspectratio]{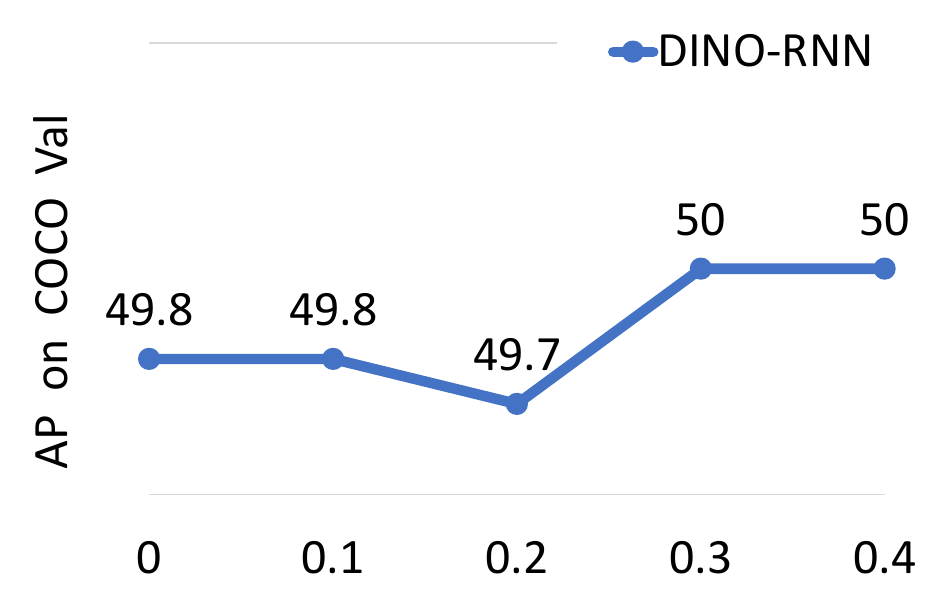}
  \subcaption{$\alpha$ sweep in Eq.~\ref{eq:sim}.}
  \label{fig:alpha_sweep_sub}
\end{subfigure}
\hfill
\begin{subfigure}[t]{0.48\linewidth}
  \centering
  \includegraphics[width=\linewidth,height=2.6cm,keepaspectratio]{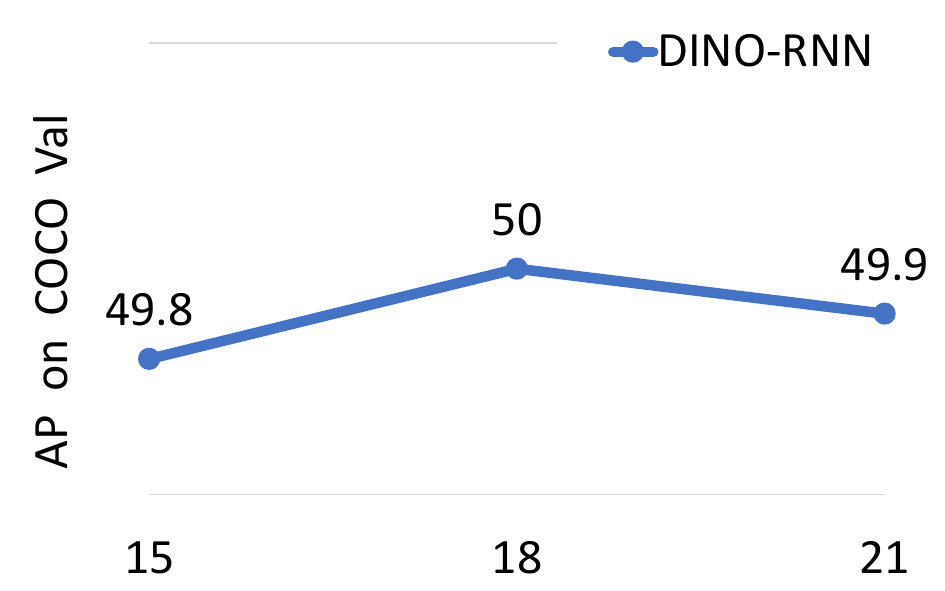}
  \subcaption{$K$ sweep in Eq.~\ref{eq:adaptive_thr}.}
  \label{fig:k_sweep_sub}
\end{subfigure}

\caption{
Sensitivity of CGQO hyperparameters on COCO val (DINO-RNN).
(a) Performance varies little across $\alpha \in [0,0.4]$.
(b) Varying the saturation constant $K$ around the default $K{=}18$ also leads to minor changes.
}
\label{fig:alpha_k_sweep}
\end{figure}

\begin{figure}[t]
    \centering
    \begin{subfigure}[t]{0.32\linewidth}
        \centering
        \includegraphics[width=\linewidth,height=2.7cm,keepaspectratio]{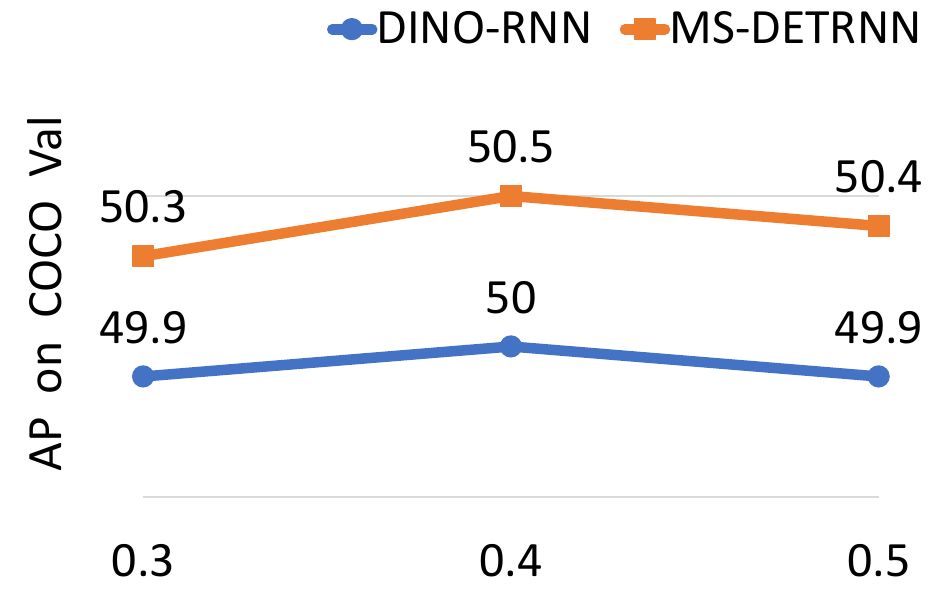}
        \subcaption{$\tau_{\min}$ sweep  ($\tau_{\max}{=}0.8$)}
        \label{fig:threshold_tmin}
    \end{subfigure}\hfill
    \begin{subfigure}[t]{0.32\linewidth}
        \centering
        \includegraphics[width=\linewidth,height=2.7cm,keepaspectratio]{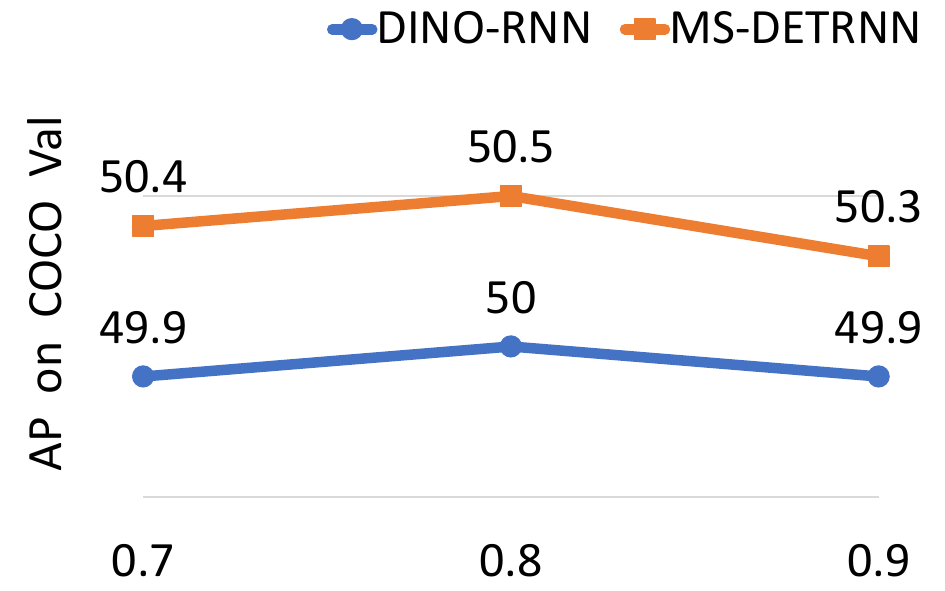}
        \subcaption{$\tau_{\max}$ sweep  ($\tau_{\min}{=}0.4$)}
        \label{fig:threshold_tmax}
    \end{subfigure}\hfill
    \begin{subfigure}[t]{0.32\linewidth}
        \centering
        \includegraphics[width=\linewidth,height=2.6cm,keepaspectratio]{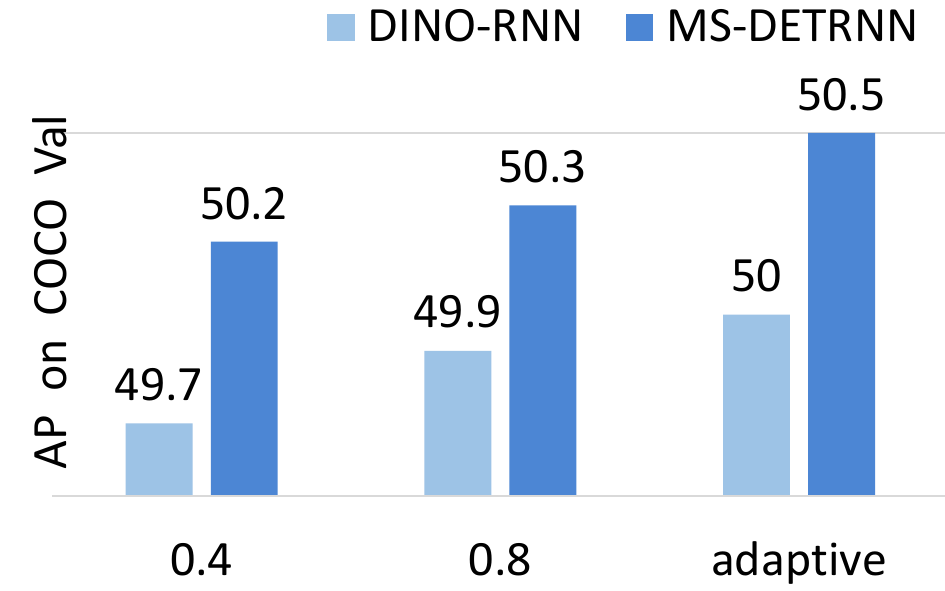}
        \subcaption{Fixed vs.\ adaptive}
        \label{fig:threshold_ad}
    \end{subfigure}

    \caption{Robustness of CGQO thresholding on COCO \texttt{val}.
(a,b) Sweeping $\tau_{\min}$/$\tau_{\max}$ (Eq.~\ref{eq:adaptive_thr}) changes AP only slightly.
(c) The density-adaptive rule performs best compared to fixed thresholds.}
\label{fig:threshold}
\end{figure}

\noindent \textbf{Default settings.}
We use the same CGQO hyperparameters for all detectors unless explicitly stated.
Specifically, we set $\alpha{=}0.3$ in Eq.~\ref{eq:sim}, and use $\tau_{\min}{=}0.4$, $\tau_{\max}{=}0.8$, and $K{=}18$ in Eq.~\ref{eq:adaptive_thr}.
Fig.~\ref{fig:alpha_k_sweep} shows that varying $\alpha$ or $K$ around these defaults only leads to minor changes, and Fig.~\ref{fig:threshold}(a,b) shows similar robustness to $\tau_{\min}/\tau_{\max}$.

\noindent \textbf{Normalization in Eq.~\ref{eq:adaptive_thr}.}
We implement $\mathrm{norm}(\cdot)$ as a fixed normalization with saturation:
\[
\mathrm{norm}(x)=\mathrm{clip}\!\left(\frac{x}{K}\right), \quad
\mathrm{clip}(x)=\min(\max(x,0),1).
\]
Here $K$ is a fixed saturation constant for the density score, preventing extremely dense neighborhoods from dominating the threshold.
As shown in Fig.~\ref{fig:alpha_k_sweep}(b), the result is stable when varying $K$ around this default.

\noindent \textbf{Robustness and Benefits of Adaptive Thresholding.}
Fig.~\ref{fig:threshold} evaluates the threshold design in CGQO on DINO-RNN and MS-DETRNN.
Sweeping the lower/upper bounds in Eq.~\ref{eq:adaptive_thr} by fixing one of $\tau_{\min}$/$\tau_{\max}$ and varying the other changes AP only slightly (within 0.1--0.2; Fig.~\ref{fig:threshold}a--b), indicating low sensitivity to these hyper-parameters.
Moreover, the density-adaptive rule consistently outperforms both fixed thresholds (0.4 and 0.8; Fig.~\ref{fig:threshold}c), suggesting a more reliable grouping criterion than a globally fixed $\tau$.

\section{Details of Bi-RNN (Mamba) module}
\label{app:iof_mamba}
\noindent \textbf{Formulation of SSMs.}
Originating from the Kalman filter, SSMs are linear time-invariant (LTI) systems that map an input $u(t) \in \mathbb{R}$ to an output $y(t) \in \mathbb{R}$ through a hidden state $\mathbf{h}(t) \in \mathbb{R}^N$. Specifically, continuous-time SSMs can be expressed by the following linear ordinary differential equations (ODEs):
\begin{equation}
\begin{aligned}
\mathbf{h}'(t) &= A\,\mathbf{h}(t) + B\,u(t),\\
y(t) &= C\,\mathbf{h}(t),
\end{aligned}
\end{equation}
where $A \in \mathbb{R}^{N \times N}$, $B \in \mathbb{R}^{N \times 1}$ and $C \in \mathbb{R}^{1 \times N}$ are the weighting parameters.

\noindent \textbf{Discretization.}
Mamba discretizes the continuous parameters $(A, B)$ into $(\bar{A}, \bar{B})$ using fixed formulas, where $(\bar{A}, \bar{B})$ is called a discretized model. Various rules can be applied, such as the zero-order hold (ZOH) in equation~(\ref{eq:zoh}), leveraging continuous-time system properties like resolution invariance and ensuring proper normalization.
\begin{equation}\label{eq:zoh}
\begin{aligned}
\bar{A} &= \exp(\Delta A),\\
\bar{B} &= (\Delta A)^{-1}\bigl(\exp(\Delta A)-I\bigr)(\Delta B).
\end{aligned}
\end{equation}
In the original Mamba \cite{Gu2023MambaLS} SSM module, the parameters $B$, $C$, and $\Delta$ are derived from linear transformations of the same input token (i.e., the object query).
In our implementation, we keep the object query as the input for generating $B$ and $C$, but \textbf{replace the input to the $\Delta$ branch} with an IOF embedding, so that the discretization step is explicitly conditioned on spatial relations.

\noindent \textbf{IOF Embedding.}
After NMS-like query sorting, given two adjacent predicted boxes \(b_i\) and \(b_{i+1}\), we compute their Intersection over Foreground (IOF) as in \eqref{eq:iof}, and encode it using the sine-cosine scheme in \eqref{eq:pe}, where \(d_p\) is the embedding dimension and $k$ is 10000. We follow DETR-style box encodings and first map the bounded IOF value to the angular range by \(2\pi\) before applying the sine-cosine embedding. The resulting embedding is linearly transformed to produce \(\Delta\).
\begin{equation}\label{eq:iof}
\text{IOF}(b_i, b_{i+1}) = \frac{\text{area}(b_i \cap b_{i+1})}{\text{area}(b_{i+1})}.
\end{equation}
\begin{equation}\label{eq:pe}
\begin{aligned}
\mathrm{Embed(IOF)}_{2j} 
&= \sin\!\left(\frac{2\pi\cdot\mathrm{IOF}}{k^{2j/d_p}}\right),\\
\mathrm{Embed(IOF)}_{2j+1} 
&= \cos\!\left(\frac{2\pi\cdot\mathrm{IOF}}{k^{2j/d_p}}\right),
\end{aligned}
\end{equation}

IOF is asymmetric and therefore provides a directional overlap cue aligned with the competition order: it measures how much a nearby competitor is covered by a stronger candidate.
This directional cue is more suitable than symmetric IoU for order-based refinement.
For bidirectional refinement, we compute the cue consistently with the processing direction: the forward pass uses \(\mathrm{IOF}(b_i,b_{i+1})\), while the reverse pass swaps the arguments and uses \(\mathrm{IOF}(b_{i+1},b_i)\).
For boundary tokens without an adjacent neighbor, we set the missing IOF value to 0 before embedding.
By injecting IOF-based embeddings into \(\Delta\), we condition the SSM discretization on adjacent spatial relations, providing an explicit order-aware spatial prior for query refinement.

\noindent \textbf{Effect of IOF Embedding.}
IOF-PE is used as a lightweight spatial conditioning in the default Mamba instantiation, while CGQO ordering and Bi-RNN refinement remain the core components.
To isolate its effect, Tab.~\ref{tab:iof_pe_ablation} removes IOF-PE while keeping CGQO and Bi-RNN unchanged.
Without IOF-PE, DETRNN still improves the attention baselines by +1.0 AP on Def-DETR++ and +0.3 AP on MS-DETR.
Adding IOF-PE further brings +0.1/+0.2 AP.
This shows that IOF-PE is a useful auxiliary spatial cue, but the main improvement does not rely on this conditioning.

\begin{table}[t]
\centering
\footnotesize
\setlength{\tabcolsep}{24pt}
\caption{
Effect of IOF positional encoding in the default Mamba instantiation on COCO val (R50, 1$\times$).
}
\label{tab:iof_pe_ablation}
\begin{tabular}{l|l|ccc}
\toprule
Detector & Variant & AP & AP$_{50}$ & AP$_{75}$ \\
\midrule
\multirow{3}{*}{Def-DETR++}
& Attn baseline & 47.6 & 65.6 & 51.7 \\
& w/o IOF-PE & 48.6 \textcolor{blue}{(+1.0)} & 66.9 & 53.1 \\
& Full & \textbf{48.7} \textcolor{blue}{(+1.1)} & \textbf{67.0} & \textbf{53.2} \\
\midrule
\multirow{3}{*}{MS-DETR}
& Attn baseline & 50.0 & 67.3 & 54.4 \\
& w/o IOF-PE & 50.3 \textcolor{blue}{(+0.3)} & 67.7 & 55.0 \\
& Full & \textbf{50.5} \textcolor{blue}{(+0.5)} & \textbf{68.2} & \textbf{55.4} \\
\bottomrule
\end{tabular}
\end{table}

\begin{table}[t]
\centering
\footnotesize
\setlength{\tabcolsep}{14pt}
\caption{
Component analysis of DETRNN on Def-DETRNN++/R50/$1\times$.
}
\label{tab:component_contributions}
\begin{tabular}{l|ccc|c}
\toprule
Variant
& CGQO + Bi-RNN
& Selected stage
& IOF-PE
& AP \\
\midrule
Baseline (self-attn)
& $\times$ & $\times$ & $\times$
& 47.6 \\
Ordered recurrent refinement
& \checkmark & $\times$ & $\times$
& 48.2 \; (+0.6) \\
+ IOF-PE
& \checkmark & $\times$ & \checkmark
& 48.4 \; (+0.8) \\
+ Selected stage
& \checkmark & \checkmark & $\times$
& 48.6 \; (+1.0) \\
Full DETRNN
& \checkmark & \checkmark & \checkmark
& \textbf{48.7 \; (+1.1)} \\
\bottomrule
\end{tabular}
\end{table}

\section{Further Component Ablation}
\label{app:further_ablation}

We further isolate the contributions of ordered recurrent refinement, IOF-PE,
and stage placement on Def-DETRNN++/R50/$1\times$ (Tab.~\ref{tab:component_contributions}).
Starting from the 47.6 AP self-attention baseline, CGQO-guided Bi-RNN reaches
48.2 AP (+0.6).
Adding IOF-PE increases the result to 48.4 AP, while using the selected
late-stage placement without IOF-PE reaches 48.6 AP.
Combining both gives the full DETRNN result of 48.7 AP (+1.1).
Thus, the main gain already comes from ordered recurrent refinement, while
IOF-PE and stage placement provide complementary improvements.

We do not treat CGQO alone as a separate interaction module because it only
constructs the query order.
As shown in Tab.~\ref{tab:ablation_ms_detr_dino}, applying CGQO without recurrent interaction
brings no clear gain, whereas Tab.~\ref{tab:ordering_variants} shows that, under the same
recurrent module, the CGQO order consistently outperforms alternative orders.
This further supports the importance of explicit competition-aware ordering.

\begin{table}[t]
\centering
\footnotesize
\setlength{\tabcolsep}{8pt}
\caption{
Profiling of query-interaction modules under different query counts $N$.
Runtime is reported in ms/image and rounded to 0.1 ms; extra peak memory is reported in MB.
CGQO+Mamba adds small runtime overhead but uses much less memory as $N$ grows, and the overhead is minor relative to the full detector cost ($\sim$50\,ms, $\sim$700\,MB at 1024$\times$1333).
}
\label{tab:module_profile}
\begin{tabular}{l|cc|cc|cc}
\toprule
\multirow{2}{*}{Module} & \multicolumn{2}{c|}{$N{=}300$} & \multicolumn{2}{c|}{$N{=}900$} & \multicolumn{2}{c}{$N{=}1500$} \\
 & Time (ms)$\downarrow$ & Mem (MB)$\downarrow$ & Time (ms)$\downarrow$ & Mem (MB)$\downarrow$ & Time (ms)$\downarrow$ & Mem (MB)$\downarrow$ \\
\midrule
Self-attn & 0.2 & 6.7  & 0.2 & 53.0  & 0.4 & 143.2 \\
Mamba     & 0.4 & 3.7  & 0.4 & 11.3  & 0.4 & 19.0 \\
CGQO      & 0.3 & 0.5  & 0.5 & 3.8   & 0.7 & 9.7 \\
\midrule
CGQO+Mamba & 0.7 \textcolor{red}{(+0.5)} & 4.2 \textcolor{blue}{(-2.5)} & 0.9 \textcolor{red}{(+0.7)} & 15.1 \textcolor{blue}{(-37.9)} & 1.1 \textcolor{red}{(+0.7)} & 28.7 \textcolor{blue}{(-114.5)} \\
\bottomrule
\end{tabular}
\end{table}

\begin{table}[t]
\centering
\footnotesize
\setlength{\tabcolsep}{7pt}
\caption{
Module-level profiling under larger query counts $N$.
Runtime is reported in ms/image and rounded to 0.1 ms; extra peak memory is reported in MB.
CGQO+Mamba greatly reduces memory compared with self-attention and has lower isolated module runtime from $N{=}4000$ onward.
}
\label{tab:module_profile_large_n}
\begin{tabular}{l|cc|cc|cc}
\toprule
\multirow{2}{*}{Module} & \multicolumn{2}{c|}{$N{=}2000$} & \multicolumn{2}{c|}{$N{=}4000$} & \multicolumn{2}{c}{$N{=}5000$} \\
 & Time (ms)$\downarrow$ & Mem (MB)$\downarrow$ & Time (ms)$\downarrow$ & Mem (MB)$\downarrow$ & Time (ms)$\downarrow$ & Mem (MB)$\downarrow$ \\
\midrule
Self-attn & 0.8 & 252.0 & 3.2 & 992.2 & 5.1 & 1545.4 \\
Mamba     & 0.4 & 25.0  & 0.4 & 49.9  & 0.4 & 62.7 \\
CGQO      & 0.9 & 16.9  & 1.9 & 65.6  & 2.3 & 101.4 \\
\midrule
CGQO+Mamba 
& 1.3 \textcolor{red}{(+0.5)} & 41.9 \textcolor{blue}{(-210.1)}
& 2.3 \textcolor{blue}{(-0.9)} & 115.5 \textcolor{blue}{(-876.7)}
& 2.7 \textcolor{blue}{(-2.4)} & 164.1 \textcolor{blue}{(-1381.3)} \\
\bottomrule
\end{tabular}
\end{table}

\section{Reproducibility of Training and Efficiency Metrics}
\label{app:efficiency_metrics}

\noindent\textbf{Training resources.}
All models are trained with the same settings as their corresponding baselines.
Unless otherwise specified, we use 8 NVIDIA RTX 4090 or 8 NVIDIA RTX 3090 GPUs with a total batch size of 16.
The standard \(1\times\) and \(2\times\) schedules correspond to 12 and 24 epochs, respectively.

\noindent\textbf{Efficiency metrics.}
All FPS measurements are conducted on a single NVIDIA RTX 4090 GPU.
We compute FLOPs with a fixed input of size \(1333\times800\times3\) (one image), using the \texttt{profile} function from the open-source THOP library.
For FPS, we measure throughput with batch size 16 on the first 160 images of COCO \texttt{val} and report the average runtime.
For module profiling in Tabs.~\ref{tab:module_profile} and~\ref{tab:module_profile_large_n}, we use the same RTX 4090 GPU, run warm-up iterations, measure CUDA time with synchronized CUDA events, and report peak allocated memory for the query-interaction module only.

\section{Profiling and Scaling}
\label{app:profiling}
\subsection{Profiling Results}
\label{app:profiling_rst}
Tabs.~\ref{tab:module_profile} and~\ref{tab:module_profile_large_n} profile the isolated runtime and extra peak memory of the main query-interaction modules as the number of queries $N$ increases.
We measure runtime with synchronized CUDA events after warm-up and round it to 0.1 ms; since individual module runtimes are small, we focus on the total CGQO+Mamba cost and the peak-memory trend.
The large-$N$ settings are used as module-level scaling stress tests, while all AP experiments use the standard query counts reported in the main tables.
Self-attention shows clear memory growth from its explicit $N{\times}N$ attention map, reaching 992.2 MB at $N{=}4000$ and 1545.4 MB at $N{=}5000$.
In contrast, CGQO+Mamba requires only 115.5 MB and 164.1 MB at the same query counts.
Its runtime is slightly slower at moderate $N$, but becomes lower in this isolated module profile from $N{=}4000$ onward, reducing time from 3.2 to 2.3 ms at $N{=}4000$ and from 5.1 to 2.7 ms at $N{=}5000$.
Since DETRNN is applied only to the last 1--2 decoder stages (Sec.~\ref{app:stage_place}), the end-to-end overhead remains small for common query counts.

\subsection{CGQO Runtime Breakdown}
\label{app:cgqo_runtime}
We profile CGQO at 900 queries using the full CUDA implementation and a
similarity-only variant with identical inputs and hyperparameters.
After 128 warm-up iterations, latency is averaged over 8,000 runs using
synchronized CUDA events.
Full CGQO takes 0.48 ms, while pairwise similarity alone takes only 0.09 ms
(about 19\%); the remaining $\sim$0.39 ms (about 81\%) comes from grouping
and order construction.

Although pairwise similarity has $O(N^2)$ complexity, it is implemented as
dense GPU operations over compact box and class-score tensors.
Most of the runtime instead comes from the subsequent grouping stage, whose
dependent control flow and irregular index updates are less GPU-friendly.

\subsection{Why CGQO+Mamba Scales}
\label{app:why_scale}
Although Algorithm~\ref{alg:cgqo} is written as a sequential greedy procedure for clarity, only the pivot selection in Lines~5--10 is inherently order-dependent.
The dense pairwise similarity \textit{all\_sim} and pivot-wise thresholds are computed once with parallel tensor operations.
For each selected pivot, Lines~7--8 read one similarity row and threshold candidate queries in parallel on GPU, and the final ordering is assembled from compact $O(N)$ grouping results.
Thus, CGQO preserves the greedy ordering semantics while keeping the expensive pairwise and per-query operations GPU-resident.

CGQO still has $O(N^2)$ pairwise similarity, but it operates on low-dimensional inputs: 4-d boxes for IoU and 80-d class scores for cosine similarity.
This gives it a much smaller constant than self-attention over 256-d query features, which forms and normalizes an $N{\times}N$ attention map.
Mamba further scales gently with $N$ due to its linear state-space computation.
Together, these design choices improve memory scaling and prevent query ordering from becoming a bottleneck at large $N$.

\begin{table}[tb]
\footnotesize
\centering
\setlength{\tabcolsep}{5.2pt}
\caption{
Full-recipe comparison with recent DETR-style SOTA on COCO val under comparable ResNet-50 / Swin-L, 1$\times$, and 900-query settings when available.
The DETRNN entries use our strongest LoRA-DETRNN full recipes.
This is a leaderboard-style context rather than a controlled module comparison, since methods use different recipes.
}
\label{tab:full_recipe_context}
\begin{tabular}{l|c|cc|cccccc}
\toprule
Method & Backbone & Epochs & Queries
& AP & AP$_{50}$ & AP$_{75}$ & AP$_S$ & AP$_M$ & AP$_L$ \\
\midrule
Def-DETR++~\cite{zhu2020deformable} & R50 & 12 & 900
& 47.6 & 65.6 & 51.7 & 30.0 & 51.0 & 61.5 \\
DINO~\cite{zhang2022dino} & R50 & 12 & 900
& 49.0 & 66.6 & 53.5 & 32.0 & 52.3 & 63.0 \\
Salience-DETR~\cite{hou2024salience} & R50 & 12 & 900
& 49.2 & 67.1 & 53.8 & 32.7 & 53.0 & 63.1 \\
DAC-DETR~\cite{hu2023dac} & R50 & 12 & 900
& 50.0 & 67.6 & 54.7 & 32.9 & 53.1 & 64.2 \\
MS-DETR~\cite{Zhao2024MSDETRED} & R50 & 12 & 900
& 50.0 & 67.3 & 54.4 & 31.6 & 53.2 & 64.0 \\
Rank-DETR~\cite{Pu2023RankDETR} & R50 & 12 & 900
& 50.4 & 67.9 & 55.2 & 33.6 & 53.8 & 64.2 \\
Stable-DETR~\cite{liu2023detection} & R50 & 12 & 900
& 50.4 & 67.4 & 55.0 & 32.9 & 54.0 & 65.5 \\
Dual-R-DETR(Route-DETR)~\cite{Zhang2025DualRDETRRQ} & R50 & 12 & 900
& 50.7 & 68.8 & 55.0 & 34.1 & 54.2 & 64.6 \\
EASE-DETR~\cite{gao2024EASEDETRET} & R50 & 12 & 900
& 50.8 & 68.9 & 55.3 & 34.1 & 54.2 & 65.1 \\
DDQ-DETR~\cite{Zhang_2023_CVPR} & R50 & 12 & 900
& 51.3 & 68.6 & 56.4 & 33.5 & 54.9 & 65.9 \\
Relation-DETR~\cite{hou2024relation} & R50 & 12 & 900
& 51.7 & 69.1 & 56.3 & 36.1 & 55.6 & 66.1 \\
Co-DETR~\cite{zong2023detrs} & R50 & 12 & 900
& 52.1 & 69.3 & 57.3 & 35.4 & 55.5 & 67.2 \\
LoRA-DETR~\cite{zhang2026integrating} & R50 & 12 & 900
& 52.5 & 69.8 & 57.3 & \textbf{37.2} & 56.5 & 66.8 \\
\rowcolor{gray!12}
DETRNN (ours) & R50 & 12 & 900
& \textbf{53.0} & \textbf{70.2} & \textbf{57.6} & 36.3 & \textbf{56.8} & \textbf{67.5} \\
\midrule
DINO~\cite{zhang2022dino} & Swin-L & 12 & 900
& 56.8 & 75.4 & 62.3 & 41.1 & 60.6 & 73.5 \\
Salience-DETR~\cite{hou2024salience} & Swin-L & 12 & 900
& 56.5 & 75.0 & 61.5 & 40.2 & 61.2 & 72.8 \\
Dual-R-DETR(Route-DETR)~\cite{Zhang2025DualRDETRRQ} & Swin-L & 12 & 900
& 57.6 & 76.4 & 63.2 & 40.8 & 61.8 & 73.6 \\
Rank-DETR~\cite{Pu2023RankDETR} & Swin-L & 12 & 900
& 57.6 & 76.0 & 63.4 & 41.6 & 61.4 & 73.8 \\
Stable-DETR~\cite{liu2023detection} & Swin-L & 12 & 900
& 57.7 & 75.7 & 63.4 & 39.8 & 62.0 & 74.7 \\
EASE-DETR~\cite{gao2024EASEDETRET} & Swin-L & 12 & 900
& 57.8 & \textbf{76.7} & 63.3 & 40.7 & 61.9 & 73.7 \\
Relation-DETR~\cite{hou2024relation} & Swin-L & 12 & 900
& 57.8 & 76.1 & 62.9 & 41.2 & 62.1 & 74.4 \\
LoRA-DETR~\cite{zhang2026integrating} & Swin-L & 12 & 900
& 58.2 & 76.2 & 63.4 & 41.8 & 63.0 & 74.4 \\
\rowcolor{gray!12}
DETRNN (ours) & Swin-L & 12 & 900
& \textbf{58.5} & 76.5 & \textbf{64.0} & \textbf{42.5} & \textbf{63.1} & \textbf{74.9} \\
\bottomrule
\end{tabular}
\end{table}

\section{Full-recipe SOTA Context}
\label{app:full_recipe_context}
Beyond the controlled comparisons in Tab.~\ref{tab:main_results_r50} and~\ref{tab:relation_swin} and the replacement study in Tab.~\ref{tab:compatibility_competition_aware}, 
Tab.~\ref{tab:full_recipe_context} compares DETRNN with recent DETR-style SOTA methods under their reported full recipes.
To reduce recipe mismatch, we focus on the common 1$\times$ training and 900-query setting with ResNet-50 and Swin-L backbones when available.
Under this leaderboard-style comparison, our DETRNN outperforms recent competition-aware methods such as EASE-DETR, Dual-R-DETR (formerly Route-DETR), DDQ-DETR, and Relation-DETR.
It also improves over strong full-recipe baselines such as Co-DETR and LoRA-DETR, achieving the highest AP among the compared DETR-style detectors under both R50 and Swin-L settings.
These results complement the controlled replacement study and show that sequential competition refinement remains effective within strong full detector recipes.

\section{Details for Competition-Aware Baselines}
\label{app:compet_aware}

Tab.~\ref{tab:compatibility_competition_aware} reports \textbf{2$\times$} results for Relation-DETR~\cite{hou2024relation}, EASE-DETR~\cite{gao2024EASEDETRET}, and DDQ-DETR~\cite{Zhang_2023_CVPR}.

\paragraph{Why replacement is used.}
Recent competition-aware DETR variants differ not only in how they model query competition, but also in many recipe-level components, such as auxiliary one-to-many supervision, dense/distinct query selection, cross-attention design, training-only branches, loss function and baseline-specific training settings.
A full best-recipe comparison would therefore mix the effect of the competition-resolution module with these additional components.
Our goal in Tab.~\ref{tab:compatibility_competition_aware} is instead to isolate the query-competition mechanism itself.
Thus, the \emph{Base} rows report the original full baselines, while the \emph{+Ours} rows keep the remaining training and detector recipe unchanged and replace only the competition-specific decoder query-interaction/pruning operation with DETRNN.
This controlled replacement setting tests whether explicit competition ordering with sequential refinement can serve as an efficient alternative to prior competition-specific modules, rather than ranking full SOTA recipes.

\paragraph{Integration rule.}
For the \emph{+Ours} rows, we remove the decoder-wise competition-specific self-attention/pruning operations and use standard self-attention in those stages, while applying DETRNN (CGQO + Bi-RNN) only at the last decoder stage.
All other settings, including the backbone, query setting, training schedule, matching/auxiliary-supervision recipe, and non-competition detector components, are kept the same as the corresponding baseline.
This is not an additive stacking comparison; it evaluates whether a lightweight last-stage ordered refinement can replace multi-stage competition-specific operations.
The results show that this replacement improves both AP and throughput, suggesting that a single ordered sequential refinement can be a more efficient alternative to multi-stage competition-specific designs.

\paragraph{Relation-DETR.}
The Base row keeps the original Relation-DETR design and training recipe.
In +Ours, we replace its competition-specific relation-attention in the decoder with DETRNN and use standard self-attention in the earlier replaced stages, while keeping the remaining detector recipe unchanged.
This reduces computation and improves AP.

\paragraph{DDQ-DETR.}
The Base row keeps DDQ-DETR(4scale)'s original dense/distinct-query design and stage-wise distinct-query selection/pruning.
In +Ours, we replace the per-stage competition-specific pruning operations with standard self-attention in earlier stages and apply DETRNN only at the last decoder, while keeping the remaining DDQ recipe unchanged.
This yields higher FPS and AP.

\paragraph{EASE-DETR.}
Since we did not find an official public implementation of EASE-DETR, we implement an EASE-Deformable-DETR++ baseline following~\cite{gao2024EASEDETRET}.
The Base rows report the corresponding full baselines, using our reproduced/evaluated implementation. The performance matches the originally reported AP.
In +Ours, we follow the same replacement rule: only the competition-specific decoder interaction is replaced by DETRNN, while the remaining EASE recipe is kept unchanged.

\section{Settings for The 1-decoder Stress Test}
\label{app:1-decoder}
The \textbf{1-decoder} results in Tab.~\ref{tab:one_decoder_nms} are obtained by training a standard 6-decoder (for both the baseline and our method) but evaluating using the predictions after the first decoder stage.
We adopt this setup because training a full 6-decoder model converges faster and is more stable than training a single-decoder from scratch, while still allowing a clean measurement of early-stage redundancy. 

Different from our default stage placement (Sec.~\ref{app:stage_place}), for this stress test we apply DETRNN at \emph{all} decoder stages so that the first-decoder output is also refined by our module.
CGQO builds the order from the shared initial predictions produced by the encoder proposal head (available in Def-DETR++, DINO, and MS-DETR).
We report AP without NMS and AP after evaluation-time NMS@0.8, where the NMS gain ($\Delta$AP) serves as a redundancy indicator.

\paragraph{Controlled protocol.}
For all baselines and DETRNN, we use the same training schedule, data processing, and loss setup, and keep the query count fixed.
At evaluation, we run the same inference and post-processing code and simply read out the predictions after the first decoder stage.
CGQO builds the order from the encoder proposal head predictions (no ground-truth information is used).

\begin{table}[tb]
\centering
\footnotesize
\setlength{\tabcolsep}{19pt}
\caption{
Stage-wise AP and AR100 of the original Def-DETR++ and MS-DETR baselines on
COCO \texttt{val2017}. AR100 saturates earlier than AP, indicating that later
decoder stages mainly refine already covered objects.
}
\label{tab:stage_ap_ar}
\begin{tabular}{l|c|ccc}
\toprule
Baseline & Metric & Stage 4 & Stage 5 & Stage 6 \\
\midrule
\multirow{2}{*}{Def-DETR++}
& AP    & 47.1 & 47.4 & 47.6 \\
& AR100 & 67.4 & 67.7 & 67.6 \\
\midrule
\multirow{2}{*}{MS-DETR}
& AP    & 49.1 & 49.5 & 50.0 \\
& AR100 & 72.7 & 73.0 & 72.6 \\
\bottomrule
\end{tabular}
\end{table}

\section{Stage-wise Analysis for Stage Placement}
\label{app:stage_analysis}

To better understand the late-stage placement of DETRNN, we examine stage-wise
AP and AR100 for the original Def-DETR++ and MS-DETR baselines.
For the original Def-DETR++ and MS-DETR baselines, Tab.~\ref{tab:stage_ap_ar}
reports stage-wise AP and AR100 from decoder stages 4--6.
In both baselines, AR100 is already close to its peak by stage 4, while AP
continues to improve through the final stage.
For Def-DETR++, AP increases from 47.1 to 47.6 from stages 4 to 6, whereas
AR100 changes only from 67.4 to 67.6.
A similar trend appears for MS-DETR, where AP increases from 49.1 to 50.0
while AR100 remains nearly saturated.

These results suggest that object coverage is largely established before the
final decoder stages, while later stages continue to refine the remaining
predictions and resolve residual query competition.
This is consistent with applying DETRNN only to the last one or two decoder
stages, as studied in Tab.~\ref{tab:stage_placement_all}.

\begin{table}[t]
\centering
\footnotesize
\setlength{\tabcolsep}{8pt}
\caption{WiderPerson consistency check (AP). NMS brings only marginal gains
(+0.1) for the baseline.}
\label{tab:widerperson_nms}
\begin{tabular}{l|ccccc}
\toprule
Detector & w/o NMS & +NMS@0.7 & +NMS@0.8 & +NMS@0.9 & +Ours \\
\midrule
Def-DETR++  & 46.2
& 46.3 (\textcolor{blue}{+0.1})
& 46.3 (\textcolor{blue}{+0.1})
& 46.3 (\textcolor{blue}{+0.1})
& \textbf{47.2} (\textcolor{blue}{+1.0}) \\
\bottomrule
\end{tabular}
\end{table}

\section{Additional Crowded Benchmark: WiderPerson}
\label{app:widerperson}
We further evaluate DETRNN on WiderPerson, a crowded pedestrian detection
benchmark with dense person instances and substantial occlusion, using the same
Def-DETR++/R50/$1\times$ setting.
As shown in Tab.~\ref{tab:widerperson_nms}, applying NMS to the baseline brings
only a marginal gain of +0.1 AP across thresholds, whereas DETRNN improves AP
from 46.2 to 47.2 (+1.0) without post-processing.
This is consistent with the CrowdHuman and COCO results in Sec.~4.5, further
showing that DETRNN can reduce duplicate competition within the decoder rather
than relying on NMS threshold tuning.

\begin{figure}[b]
\centering
\hspace{-0.2\textwidth}
\begin{minipage}[c]{0.4\textwidth}
  \centering
  \includegraphics[width=0.62\textwidth]{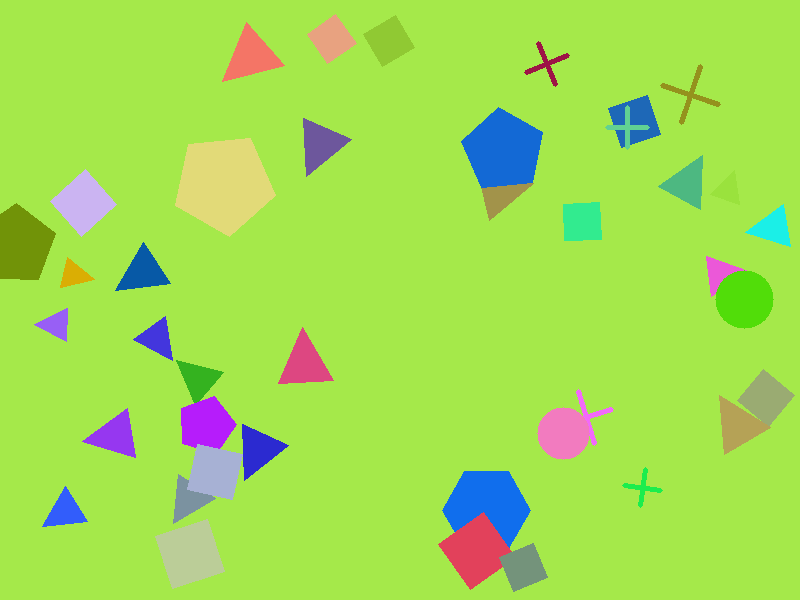}
  \subcaption{Synthetic data example.} \label{fig:syn_data}
\end{minipage}
\hspace{-0.12\textwidth}
\begin{minipage}[c]{0.4\textwidth}
  \centering
  \includegraphics[width=0.75\textwidth]{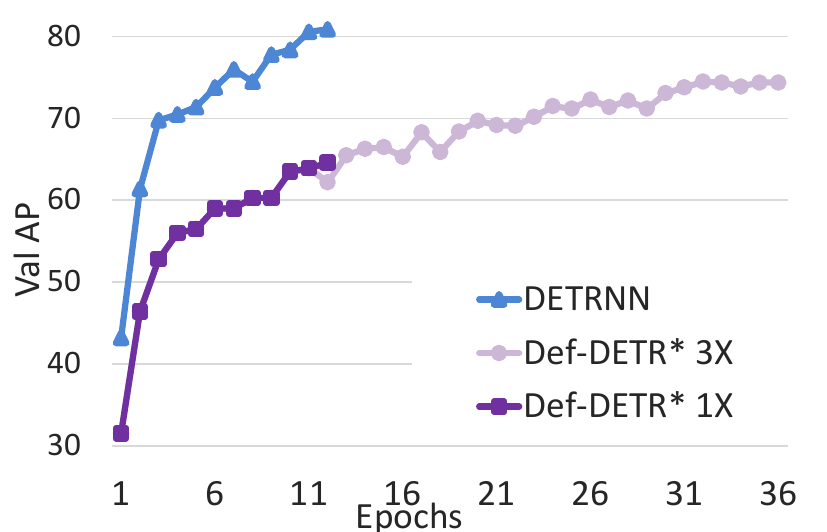}
  \subcaption{Convergence comparison.} \label{fig:conv_comparison}
\end{minipage}
\hspace{-0.05\textwidth}
\begin{minipage}[c]{0.32\textwidth}\scriptsize
  \centering
  \setlength{\tabcolsep}{4pt}
  \begin{tabular}{l|c|c|c|c}
  \toprule
  Method & Epochs & NMS & AP & FPS \\
  \midrule
  \multirow{2}{*}{Def-DETR* } & \multirow{2}{*}{12} &  & 64.6 & 124  \\
  &  & {\checkmark} & 81.9 (+17.3) &  118 \\
  \midrule
  \multirow{2}{*}{Def-DETR* } & \multirow{2}{*}{36} &  & 74.4 &  124 \\
  &  & {\checkmark} & 83.3 (+8.9) & 118 \\
  \midrule
  \multirow{2}{*}{DETRNN (ours)} & \multirow{2}{*}{12} &  & 83.6 & 120 \\
  &  & {\checkmark} & 83.2 (-0.4) & 114 \\
  \bottomrule
  \end{tabular}
  \subcaption{Synthetic experiment results.
  } \label{table:synthetic}
\end{minipage}

\caption{(a), (b), and (c) show the synthetic data example, convergence comparison on the synthetic val dataset, and experiment results comparing Def-DETR* \cite{zhu2020deformable} and DETRNN with and without NMS, respectively. The dataset and model are simplified for rapid validation.
}
\label{fig:synthetic_results}
\end{figure}

\section{Synthetic Mechanism Validation}
\label{subsec:synthetic}
To visualize the effect of competition-aware ordering in a controlled setting, we construct a synthetic benchmark (Fig.~\ref{fig:syn_data}) that intentionally amplifies object redundancy (dense overlaps) and compare DETRNN with a simplified version of Def-DETR~\cite{zhu2020deformable} (named Def-DETR$^\ast$) under the same setup.
As shown in Fig.~\ref{fig:conv_comparison} and Fig.~\ref{table:synthetic}, DETRNN reaches 83.6 AP within 12 epochs, while Def-DETR$^\ast$ attains 64.6 AP at 12 epochs and 74.4 AP at 36 epochs.
Moreover, since NMS mainly removes duplicates, the AP change after applying it reflects how many redundant predictions remain.
This change is large and positive for Def-DETR$^\ast$, but near zero or slightly negative for DETRNN, indicating that DETRNN leaves few duplicates for NMS to remove.
For DETRNN, the slight AP drop suggests that once redundancy is largely suppressed, hard thresholding may start removing useful boxes.
We include this synthetic benchmark only as a fast, controlled sanity check; the main conclusions are still based on MS-COCO and CrowdHuman.

\subsection{Settings}
\label{app:syn_set}
We conduct a synthetic experiment on a dataset of 4800 images, each sized $800\times600$. Every image features a random background and foreground objects drawn from six shape categories (triangle, square, pentagon, hexagon, circle, and cross) in random colors, positions, and sizes, with the number of objects per image varying from 1 to 40 and each object occupying no more than one-tenth of the image size. Our model is a simplified version of Def-DETR~\cite{zhu2020deformable} with a frozen R50 backbone, a single encoder, and a three-level feature pyramid paired with a single decoder. We further reduce the decoder hidden dimensions to one quarter of the original setting, use 300 object queries, and train with batch size 16. The threshold in CGQO is a simple fixed 0.4. Owing to the small image size, limited dataset, and simplified model, the synthetic experiment runs very fast, making it ideal for rapid hypothesis validation.

\section{Cross-Group Transitions in the Recurrent Sequence}
\label{app:group_boundary}

CGQO concatenates multiple competition groups into a single recurrent sequence.
We therefore examine whether transitions between successive groups are
distinguishable in the query features.

Using Def-DETRNN++/R50/$1\times$ on the first 1,600 COCO
\texttt{val2017} images, we refer to the last query of a group as its
\textit{tail}, and the preceding query as \textit{tail-1}.
We compare the cosine similarity of the last within-group pair
(\textit{tail-1} vs.\ \textit{tail}) with that of the pair crossing the group
boundary (\textit{tail} vs.\ \textit{next pivot}).

As shown in Tab.~\ref{tab:group_boundary}, the mean similarity drops from
0.734 within a group to 0.330 across the boundary.
Moreover, the within-group 25th percentile (0.610) remains higher than the
boundary 75th percentile (0.486), indicating a clear feature transition
between successive competition groups.

\begin{table}[t]
\centering
\footnotesize
\setlength{\tabcolsep}{12pt}
\caption{Cosine similarity around CGQO group boundaries.}
\label{tab:group_boundary}
\begin{tabular}{l|ccc}
\toprule
Compared queries & Mean & 25th percentile & 75th percentile \\
\midrule
\textit{tail-1} vs.\ \textit{tail}
& \textbf{0.734} & \textbf{0.610} & \textbf{0.906} \\
\textit{tail} vs.\ \textit{next pivot}
& 0.330 & 0.133 & 0.486 \\
\bottomrule
\end{tabular}
\end{table}

\begin{table}[b]
\centering
\footnotesize
\setlength{\tabcolsep}{5pt}
\caption{Multi-seed results on COCO. Mean and standard deviation are computed
over three runs.}
\label{tab:seed_variance}
\begin{tabular}{l|c|ccc|cc}
\toprule
Detector & Baseline & Seed 1 & Seed 2 & Seed 3 & Mean & Std. \\
\midrule
Def-DETRNN++ & 47.6 & 48.7 & 48.9 & 48.6 & 48.73 & 0.15 \\
DINO-RNN     & 49.0 & 50.0 & 50.0 & 49.8 & 49.93 & 0.12 \\
MS-DETRNN    & 50.0 & 50.5 & 50.3 & 50.5 & 50.43 & 0.12 \\
LoRA-DETRNN  & 52.5 & 53.0 & 53.0 & 52.9 & 52.97 & 0.06 \\
\bottomrule
\end{tabular}
\end{table}

\section{Seed Variance and Reliability}
To assess run-to-run variance, we repeat four representative R50/$1\times$
settings with three random seeds. Results are summarized in
Tab.~\ref{tab:seed_variance}.

All runs outperform their corresponding baselines, with standard deviations of
only 0.06--0.15 AP. This indicates that the improvements are consistent across
random seeds and are substantially larger than the observed run-to-run variation.

\begin{table}[t]
\centering
\footnotesize
\setlength{\tabcolsep}{9pt}
\caption{Comparison of order-aware interaction modules with CGQO on
DINO/R50/$1\times$.}
\label{tab:interaction_modules}
\begin{tabular}{l|c|ccc}
\toprule
Method & Params. & AP & AP$_{50}$ & AP$_{75}$ \\
\midrule
Self-attention baseline & 47.5M & 49.0 & 66.6 & 53.5 \\
\midrule
CGQO + Uni-RNN          & 47.6M & 49.5 & 67.7 & 53.9 \\
CGQO + rank-bias attn   & 47.7M & 49.7 & 67.8 & \textbf{54.1} \\
CGQO + causal attn      & 47.5M & 49.5 & 67.6 & 54.0 \\
CGQO + 1D Conv          & 47.7M & 49.6 & 67.6 & \textbf{54.1} \\
CGQO + Bi-RNN           & 47.6M & \textbf{50.0} & \textbf{68.1} & 54.0 \\
\bottomrule
\end{tabular}
\end{table}

\section{Alternative Order-Aware Interaction Modules}
\label{app:interaction_modules}
To examine how broadly the CGQO order can be exploited, we compare Bi-RNN with
several alternative order-aware interaction mechanisms on DINO/R50/$1\times$,
while keeping CGQO fixed.
Uni-RNN processes the ordered queries in one direction; rank-bias attention
injects their relative CGQO ranks into self-attention as bias; causal attention
restricts attention according to the query order; and 1D convolution aggregates
local neighbors along the ordered sequence.

All four alternatives improve over the self-attention baseline, showing that
the CGQO order can benefit different order-aware interaction mechanisms.
Bi-RNN achieves the highest overall AP and AP$_{50}$ with a comparable parameter
count, supporting its use as our default design.

\section{Limitations, Future Directions, and Broader Impact}
\label{app:limitations}

\paragraph{Limitations.}
CGQO uses a discrete, non-differentiable, hand-designed grouping rule rather
than a learned ordering objective, and therefore depends on the quality of
preceding predictions.
DETRNN is designed for late-stage refinement, and its gains can be smaller on
already strong or competition-aware baselines.
Our evaluation does not cover some recent detector families such as RT-DETR and
D-FINE, while extreme cases including very large or chain-structured competition
groups and unusual query distributions remain insufficiently explored.

\paragraph{Future directions.}
CGQO is not tied to box IoU and only requires a pairwise similarity measure,
allowing extensions such as mask IoU for instance segmentation or task-specific
similarities for 3D and video detection.
Its favorable large-query memory and runtime scaling also motivates further
study in dense detection settings such as remote-sensing imagery.

\paragraph{Broader impact.}
DETRNN may benefit applications such as autonomous perception and assistive
vision, while also inheriting general object-detection risks including
privacy-sensitive surveillance and biased deployment.

\section{Visualization of prediction redundancy}

\begin{figure}[t]
    %
    
    \includegraphics[width=14.cm]{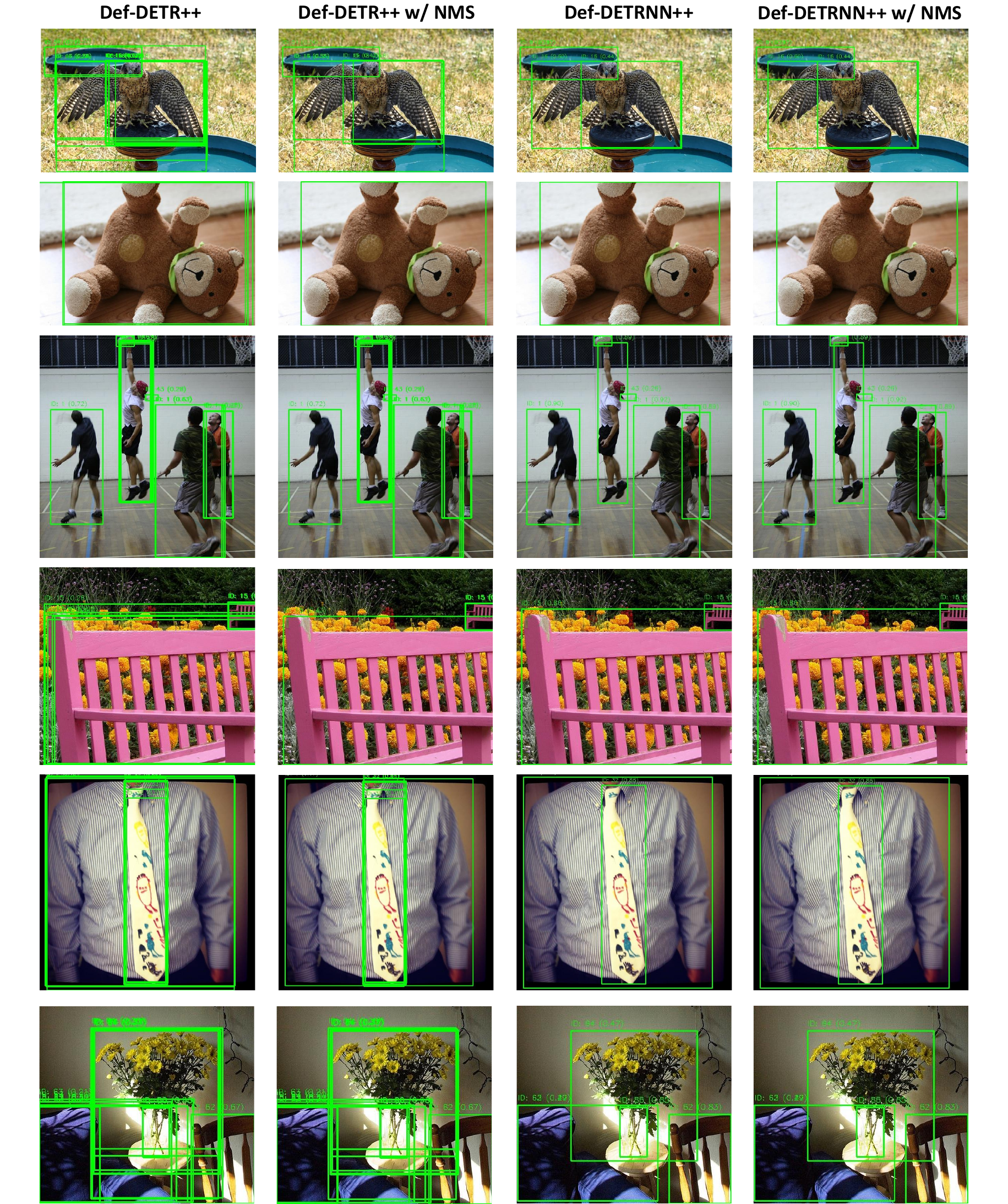}
    \hspace{-1cm}
    \caption{Visual comparison of predictions for Def-DETR++ and Def-DETRNN++, with and without NMS, under the 1-decoder setting. The results highlight Def-DETRNN++'s minimal redundancy and effective end-to-end predictions, while Def-DETR++ shows significant redundancy without NMS.
    }
    \label{fig:vis_nms}
\end{figure}

The visual comparison in Fig.~\ref{fig:vis_nms} highlights the differences in redundancy handling between the models. Without NMS, the baseline predictions contain many overlapping duplicates.
Even after applying NMS with $\tau{=}0.7$ (score threshold $1e{-}1$), noticeable duplicates remain, indicating limited redundancy suppression inside the decoder. In contrast, Def-DETRNN++ predictions show minimal redundancy, whether or not NMS is applied, demonstrating its capability for substantially less redundant end-to-end predictions. This visualization underscores the advantages of Def-DETRNN++ in accelerating convergence and achieving effective query differentiation, especially in the 1-decoder setting.


\clearpage

\end{document}